\documentclass{article}

\PassOptionsToPackage{numbers,compress}{natbib}

\usepackage[preprint]{neurips_2026}

\usepackage[utf8]{inputenc}
\usepackage[T1]{fontenc}
\usepackage{hyperref}
\usepackage{url}
\usepackage{booktabs}
\usepackage{amsfonts}
\usepackage{amsmath}
\usepackage{amssymb}
\usepackage{mathtools}
\usepackage{microtype}
\usepackage{xcolor}
\usepackage{array}
\usepackage{multirow}
\usepackage{graphicx}
\usepackage{algorithm}
\usepackage{algpseudocode}
\usepackage[section]{placeins}

\title{U-HNO: A U-shaped Hybrid Neural Operator with Sparse-Point Adaptive Routing for Non-stationary PDE Dynamics}

\author{%
  Yingzhe Ma$^{1}$ \quad Xiao Yang$^{1}$ \quad Yuxin Xie$^{2}$ \quad Zihan Xiong$^{1}$ \quad Jinliang Liu$^{1}$ \\
  $^{1}$University of Electronic Science and Technology of China \quad
  $^{2}$Peking University
}

\newcommand{\reals}{\mathbb{R}}
\newcommand{\norm}[1]{\left\lVert #1 \right\rVert}
\newcommand{\abs}[1]{\left\lvert #1 \right\rvert}
\newcommand{\fwd}{\mathcal{F}}
\newcommand{\ifwd}{\mathcal{F}^{-1}}

\begin{document}

\maketitle

\begin{abstract}
Solutions to many partial differential equations (PDEs) display
coexisting smooth global transport and localized sharp features
within a single trajectory: shock fronts, thin interfaces, and
concentrated high-frequency content sit on top of slowly varying
backgrounds. This poses a fundamental challenge for neural operators:
Fourier-based architectures mix nonlocal interactions efficiently
through frequency-domain parameterization but tend to under-resolve
localized non-smooth features, whereas spatially local architectures
recover fine detail at the cost of long-range propagation and
rollout stability.
Existing hybrid operators paper over this tension with a fixed,
spatially uniform fusion---additive concatenation or a single global
gate---that forces the same trade-off everywhere in the field.

We propose U-HNO, a U-shaped hybrid neural operator whose
central design is Sparse-Point Adaptive Routing (SPAR):
at every spatial location, a per-pixel hard mask selects whether
the global Fourier branch or the local multi-scale Gaussian branch
should dominate, and the sparsity ratio itself is a function of
the local contrast of the routing signal, so smooth regions and
shock-aligned regions receive different mixtures of global and
local computation. SPAR is embedded in a hierarchical
encoder--bottleneck--decoder backbone with skip connections so that
the dual branches and the gate operate at every resolution. Training combines pointwise supervision with
a finite-difference $H^1$ gradient term and a band-wise spectral
consistency regularizer, jointly targeting field accuracy, gradient
fidelity, and frequency stability across rollout horizons.

Across a heterogeneous benchmark suite spanning 1D Burgers,
Kuramoto--Sivashinsky, and KdV equations, 2D advection,
Allen--Cahn, Navier--Stokes, and Darcy flow, as well as 3D
compressible Navier--Stokes at transonic Mach number from PDEBench,
U-HNO achieves state-of-the-art
rollout accuracy on the majority of tasks in both relative $L^2$
and gradient-aware $H^1$ metrics, with the largest gains on
problems dominated by sharp localized features such as 1D Burgers
and the transonic 3D case. An ablation across nine
variants shows that removing any single component---the hierarchical
backbone, the dual-branch design, the contrast-adaptive SPAR gate,
or the structured loss---substantially degrades rollout error,
indicating that the gains do not concentrate in any one ingredient.
\end{abstract}

\section{Introduction}
\label{sec:intro}

Operator learning approximates mappings between infinite-dimensional
function spaces and provides fast surrogates for repeated PDE
simulation, inverse problems, and design \citep{li2020fourier,kochkov2021machine}.
Fourier neural operators (FNO) \citep{li2020fourier} are popular because
frequency-domain parameterization mixes nonlocal interactions
efficiently and generalizes across resolutions. Their weakness is the
flip side of this strength: PDE solutions are rarely spectrally
uniform. Smooth large-scale transport coexists with sharp transitions,
boundary layers, shocks, and fine interfaces. Networks are biased
toward low frequencies \citep{rahaman2019spectral}, and a flat global
mode budget allocates the same spectral capacity everywhere
irrespective of local frequency content. Spatially local operators
\citep{li2020neural,pfaff2020learning,brandstetter2022message} attack the
bias from the opposite side, improving local detail at the cost of
efficient long-range propagation. The gap is a neural operator that
preserves \emph{both} local spatial fidelity and global spectral
coherence, which is most acute for non-stationary spectral fields where
the dominant frequency varies in space, time, and initial condition.

\textbf{Approach.} We introduce \textbf{U-HNO}. A multi-scale
Gaussian local branch and a Fourier branch run in parallel inside a
U-shaped backbone with skip connections \citep{rahman2022u}, exposing
the hybrid at every resolution. A Sparse-Point Adaptive Routing (SPAR)
gate, drawing on content-adaptive sparse sampling
\citep{kirillov2020pointrend} and sparsely-gated MoEs
\citep{shazeer2017outrageously}, emits a per-pixel hard choice between
the two branch outputs in the forward pass and masks the dominant
output-gradient at that location to the chosen branch in the backward
pass; both branches are evaluated everywhere, so SPAR is a
representational dispatcher rather than a compute-saving mechanism.
Training combines single-step MSE, a finite-difference gradient term,
and a cross-branch consistency regularizer.

\begin{figure}[!htbp]
  \centering
  \includegraphics[width=\linewidth]{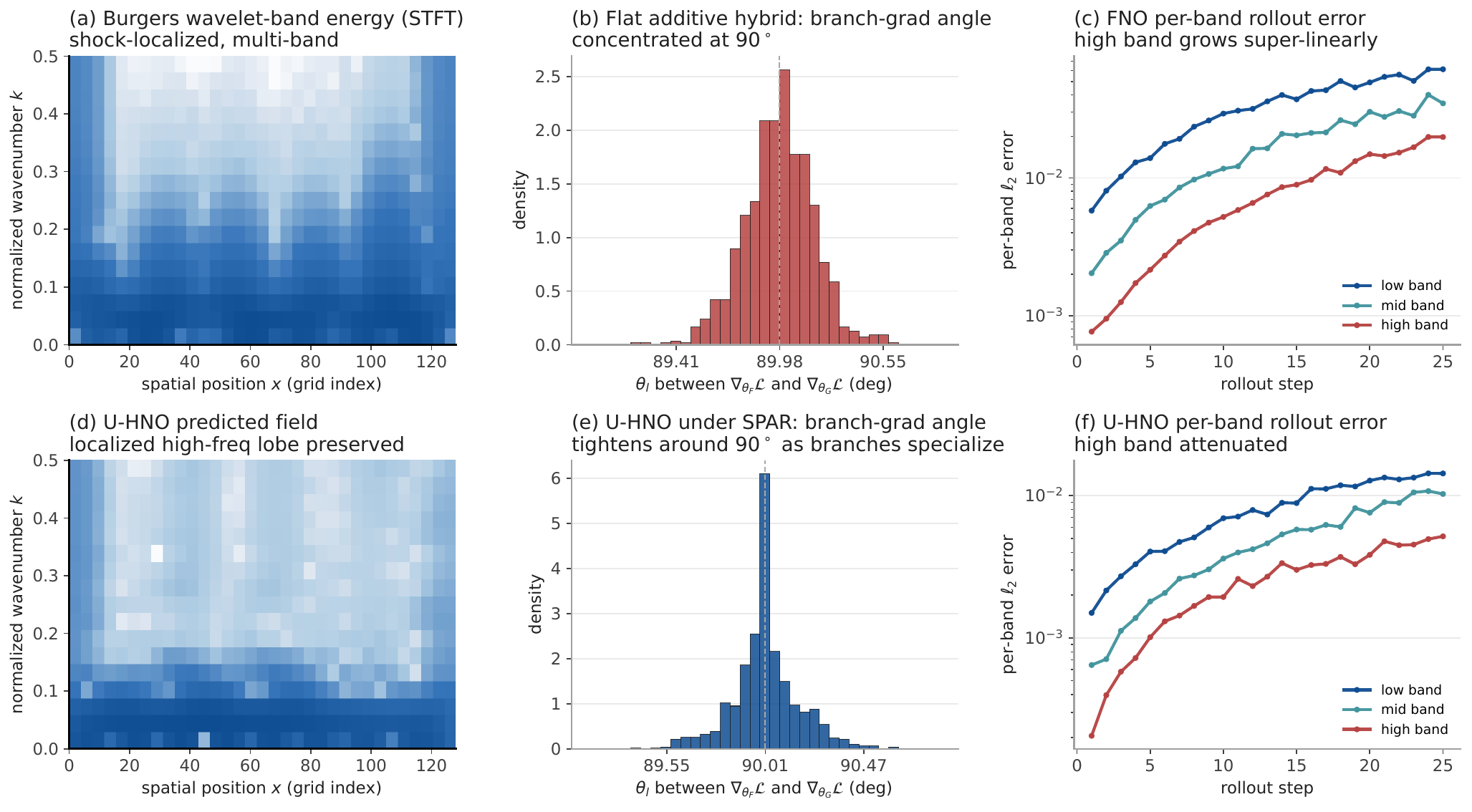}
  \caption{\textbf{Three diagnostics motivating U-HNO.}
  (a) Burgers wavelet-band energy, localized near shocks;
  (b) flat-additive branch-grad angle peaks at $90^{\circ}$ on NS
  (App.~\ref{app:angle}); (c) FNO per-band $\ell_2$ on Kolmogorov flow
  with super-linear high-band growth.
  (d--f) U-HNO mirror panels with all three pathologies attenuated.}
  \label{fig:motivation}
\end{figure}

\textbf{Why this inductive bias.} Figure~\ref{fig:motivation}
visualizes three diagnostics. (a) Frequency content varies sharply
within a single Burgers trajectory---mid/high-frequency energy
concentrates near shocks while ambient regions stay low-frequency, so
uniform spectral truncation must trade off one regime against another.
(b) Inside a flat additive FNO+local hybrid, the layer-wise cosine
angle between $\nabla_{\theta_F}\mathcal{L}$ and
$\nabla_{\theta_G}\mathcal{L}$ on NS (App.~\ref{app:angle})
concentrates at $90^{\circ}$, indicating near-orthogonal training
signals. (c) Pure spectral baselines accumulate high-frequency drift
under long rollout that pointwise MSE does not suppress.
Panels (d--f) show U-HNO mirror diagnostics on the same tasks: all
three pathologies are attenuated. These motivate (i) a multi-resolution
U-shape, (ii) per-pixel commitment of SPAR over additive fusion, and
(iii) a structured loss penalizing gradient and cross-branch divergence.

\textbf{Contributions.}
(i) Three diagnostics on Burgers/NS documenting non-stationary
spectral structure as a failure mode: shock-localized band energy,
near-orthogonal branch-grads in flat additive hybrids, and super-linear
high-band rollout drift in pure spectral baselines.
(ii) \textbf{U-HNO}, a U-shaped Gaussian--Fourier hybrid with
\emph{Sparse-Point Adaptive Routing} (SPAR)---a per-pixel hard MUX
with contrast-adaptive keep-ratio---and an asymmetric decoder routing
the high-res skip to the local branch and the upsampled coarse
context to the Fourier branch.
(iii) A single-step structured objective combining MSE, a
finite-difference $H^1$ term, and a cross-branch consistency (CBC)
regularizer; no rollout supervision required.
(iv) On eight 1D/2D/3D PDE benchmarks including 3D transonic
compressible Navier--Stokes \citep{NEURIPS2022_0a974713},
U-HNO improves the joint $(\operatorname{relL2},\operatorname{relH1})$
trade-off over seven baselines; a nine-mode ablation isolates each
component (largest $\operatorname{relH1}$ swing $4.1{\times}$ from
local-branch removal).

\section{Related Work}
\label{sec:related}

\paragraph{Neural operators.}
DeepONet \citep{lu2021learning} and the neural operator framework
\citep{kovachki2023neural} establish learning maps between function
spaces; FNO \citep{li2020fourier} parameterizes integral kernels in the
Fourier domain. Subsequent spectral work covers factorized/adaptive
mode allocation \citep{tran2021factorized,guibas2021adaptive,kalimuthu2025loglo},
geometry-aware variants \citep{li2023fourier,bonev2023spherical},
wavelet bases \citep{tripura2022wavelet}, and Galerkin/Fourier
transformers \citep{cao2021choose,hao2023gnot,wu2024transolver}. The
mode budget is set globally and cannot adapt to fields whose dominant
frequency varies in space and time, and fixed-truncation mixing
\citep{rahaman2019spectral} under-represents high-frequency local
content.

\paragraph{Local and hybrid operators.}
Graph and message-passing solvers
\citep{li2020neural,pfaff2020learning,brandstetter2022message} and
multigrid CNNs \citep{he2019mgnet} model short-range interactions
explicitly. Among hybrids: \emph{U-NO} \citep{rahman2022u} adapted the U-Net
encoder--decoder of \citet{ronneberger2015u} to operator learning but
instantiates the same spectral mixer at every scale;
\emph{Conv-FNO} \citep{liu2025enhancing} adds a parallel local conv via
\emph{additive} fusion in a flat stack; \emph{LogLo-FNO}
\citep{kalimuthu2025loglo} reweights the Fourier mode budget rather
than adding a local branch; \emph{CNO} \citep{raonic2023convolutional} composes
heterogeneous basis blocks without per-pixel routing. U-HNO inherits
the U-backbone but adds a multi-scale Gaussian branch at every level,
makes the decoder asymmetric (local consumes skip, spectral consumes
upsampled bottleneck), and replaces additive fusion with a per-pixel
hard MUX.

\paragraph{Sparse routing and structured losses.}
SPAR draws on sparsely-gated MoE \citep{shazeer2017outrageously} and
content-adaptive sparse sampling in dense prediction
\citep{kirillov2020pointrend}, but unlike MoE it computes both
branches everywhere and uses sparsity only to select \emph{which
branch's output is emitted} and \emph{which receives the dominant
output-gradient} at each pixel; SPAR addresses representational
specialization rather than a compute budget. The training objective
combines MSE with a finite-difference $H^1$ term in the spirit of
PINO \citep{li2024physics} and PINNs \citep{raissi2019physics}, plus
a feature-space cross-branch consistency regularizer (closer to
multi-branch consistency in dense prediction than to band-wise
spectral matching \citep{brandstetter2022message}).

\section{Method}
\label{sec:method}

\paragraph{Setup.}
Let $G^\star: \mathcal{A} \to \mathcal{U}$ be a target operator on a
spatial domain $\Omega$. Supervised operator learning seeks $G_\theta$
minimizing $\mathbb{E}_{a\sim\mu}\,\mathcal{L}(G_\theta(a),G^\star(a))$
under input distribution $\mu$. For evolutionary PDEs we additionally
consider a single-step transition $T_\theta(\hat u_t,c)\mapsto\hat
u_{t+1}$ applied autoregressively from $u_0$. We say a field has
\emph{non-stationary spectral structure} when its windowed spectrum
$\hat u_{\mathrm{loc}}(x;\xi)$ is highly non-uniform across $x$.

\begin{figure}[!htbp]
  \centering
  \includegraphics[width=\linewidth]{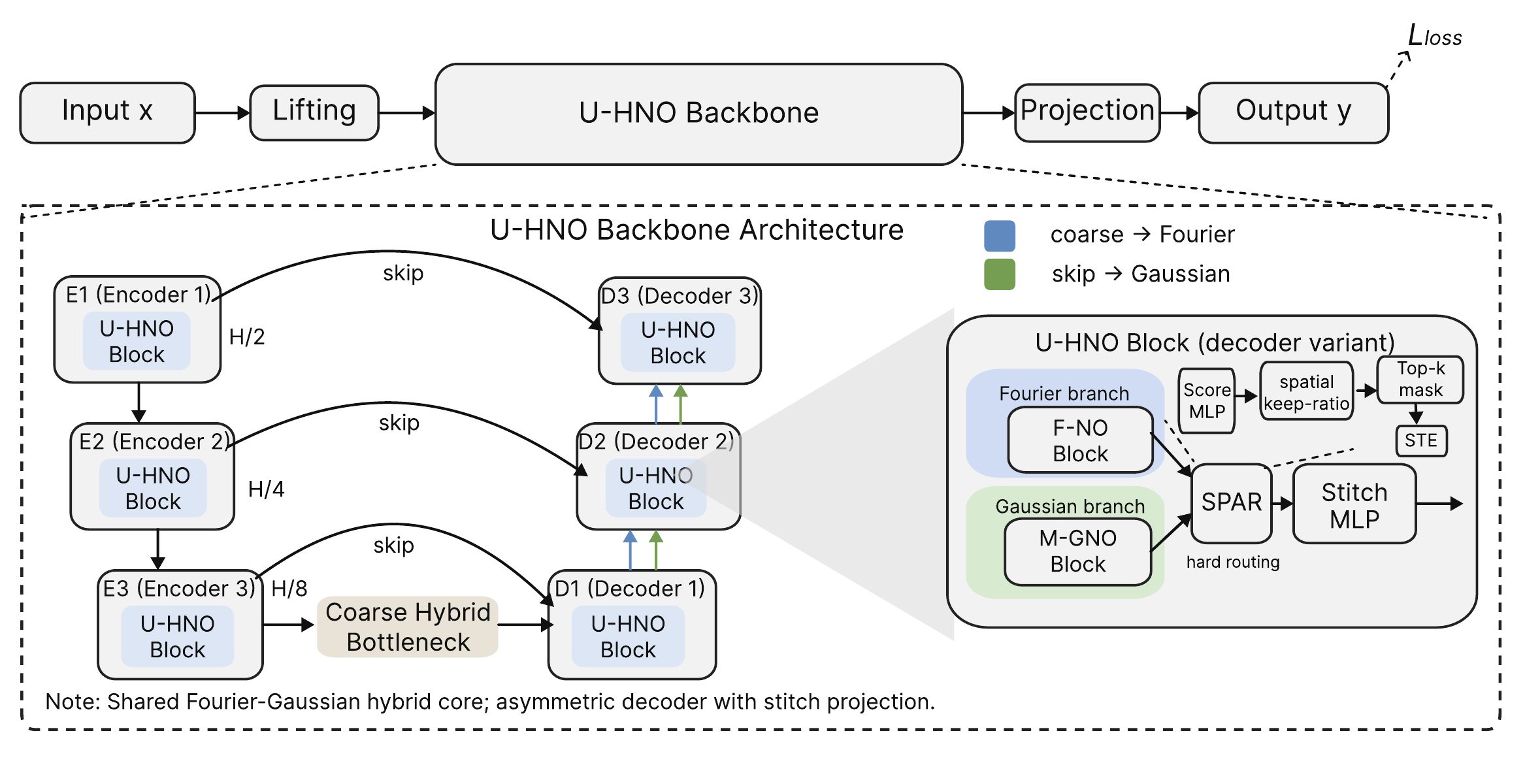}
  \caption{\textbf{U-HNO framework.}
  Lifting--Backbone--Projection pipeline; $L$-level U-shape backbone
  ($H/2,H/4,H/8$). The decoder is \emph{asymmetric}: upsampled coarse
  features (blue) feed the Fourier branch $\mathcal{B}_F$, while
  encoder skips (green) feed the multi-scale Gaussian branch
  $\mathcal{B}_G$. Both branches run in parallel; a Score MLP on
  $[z_F;z_G]$ drives the contrast-adaptive spatial keep-ratio
  $\rho^{\ell}$ (Eq.~\eqref{eq:spar-rho}), and SPAR emits a Top-$k$
  hard mask optimized via a straight-through estimator (STE). The
  routed feature is stitched and projected to the next level.}
  \label{fig:framework}
\end{figure}

\subsection{U-HNO Block}
\label{sec:method-block}
At level $\ell$ both branches are evaluated for every spatial location,
\begin{equation}
z_F^{\ell}=\mathcal{B}_F(h^{\ell}),\qquad
z_G^{\ell}=\mathcal{B}_G(h^{\ell}).
\end{equation}
SPAR (Sec.~\ref{sec:method-spar}) emits a per-pixel hard mask
$g^{\ell}(x)\in\{0,1\}$ and outputs
$r^{\ell}(x)=g^{\ell}(x)\,z_F^{\ell}(x)+(1{-}g^{\ell}(x))\,z_G^{\ell}(x)$.
A residual $1\!\times\!1$ projection $W_R^{\ell}$ on $h^{\ell}$ is
added and the result passed through GELU,
\begin{equation}
h^{\ell+1}=\mathrm{GELU}\bigl(W_R^{\ell}h^{\ell}+r^{\ell}\bigr).
\label{eq:block}
\end{equation}
Two design properties distinguish a U-HNO block from a flat additive
hybrid: \emph{(i)} fusion is a per-pixel hard MUX rather than a
weighted concatenation, so each pixel commits to exactly one
representation and the two parameter sets specialize on disjoint
spatial regions rather than fighting for the same locations
(both branches still compute everywhere); \emph{(ii)} the block is
wrapped inside a U-shape (Sec.~\ref{sec:method-ushape}), so the same
hybrid runs at every resolution and each branch sees features at the
scale where it is most informative.

\subsection{U-shaped Backbone with Hybrid Branches}
\label{sec:method-ushape}
The block is wrapped in a U-shape with $L$ levels \citep{rahman2022u};
encoder steps apply a U-HNO block then $3{\times}3$ stride-2 down to
$C_\ell=C_0\cdot 2^{\ell}$ channels, with a single block at the
coarsest grid as the bottleneck. The decoder is \emph{asymmetric}:
\begin{equation}
\tilde h^{\ell-1}=W_{\mathrm{up}}^{\ell}\,\mathrm{Up}(h^{\ell}),\quad
z_F^{\ell-1}=\mathcal{B}_F(\tilde h^{\ell-1})+W_R^{\ell-1}\tilde h^{\ell-1},\quad
z_G^{\ell-1}=\mathcal{B}_G(e^{\ell-1}),
\label{eq:dec-asym}
\end{equation}
where $e^{\ell-1}$ is the encoder skip and $\mathrm{Up}$ is bilinear
upsampling: the spectral branch integrates upsampled coarse global
flow while the Gaussian branch refines high-resolution skip detail
(see App.~\ref{app:asym-dec}). The retained-mode cap
$K^{\mathrm{eff}}_{\ell}=\min(K_0,\lfloor N_{\ell}/2\rfloor)$
(Eq.~\eqref{eq:mode-cap}) discards modes above the down-sampled
Nyquist, so coarser levels carry proportionally fewer modes at a
uniform parameter budget; we do not claim a Galerkin-style guarantee.
\begin{equation}
K^{\mathrm{eff}}_{\ell}=\min\bigl(K_0,\,\lfloor N_{\ell}/2\rfloor\bigr).
\label{eq:mode-cap}
\end{equation}

The Fourier branch is the standard spectral conv
\citep{li2020fourier}, $z_F^{\ell}=\ifwd(R_\theta^{\ell}\odot\fwd(h^{\ell}))$
with $R_\theta^{\ell}\in\mathbb{C}^{C_\ell\times C_\ell\times K_0\times K_0}$
indexed at $K^{\mathrm{eff}}_{\ell}$ at runtime; conjugate symmetry
is preserved on the rfft axis (App.~\ref{app:fourier-detail}). The
multi-scale Gaussian branch stacks $M{=}3$ \emph{normalized} depthwise
Gaussian convolutions with learnable scales
$\log\sigma_m\in\reals^G$ (init $\{0.5,1.0,2.5\}$),
\begin{equation}
k_m(\Delta;\sigma_m)=
\frac{\exp(-\norm{\Delta}^2/2\sigma_m^2)}
{\sum_{\Delta'\in\Omega_m}\exp(-\norm{\Delta'}^2/2\sigma_m^2)},\quad
\Omega_m=\{\Delta:\norm{\Delta}_\infty\le\lceil 3\sigma_m\rceil\},
\label{eq:gaussian-kernel}
\end{equation}
applied as $z_m^{\ell}=W_m^{\mathrm{post}}(k_m\star_{\mathrm{circ}}(W_m^{\mathrm{pre}}h^{\ell}))$
and fused by a $1{\times}1$ conv $+$ GELU; per-kernel normalization
decouples amplitude from $\sigma_m$ so routing sees unbiased magnitudes
(App.~\ref{app:gaussian-impl}).

\subsection{Sparse-Point Adaptive Routing (SPAR)}
\label{sec:method-spar}
\paragraph{Routing logits.}
A two-layer pointwise MLP on the channel-concatenated branch outputs
produces a scalar logit per pixel:
\begin{equation}
s^{\ell}(x)=W_2^s\,\mathrm{GELU}\bigl(W_1^s\,[z_F^{\ell}(x);\,z_G^{\ell}(x)]\bigr),\qquad s^{\ell}(x)\in\reals.
\label{eq:spar-score}
\end{equation}

\paragraph{Contrast-adaptive keep-ratio.}
Let $\bar s^{\ell}=\mathbb{E}_x\abs{s^{\ell}(x)}$ and
$\sigma_s^{\ell}=\mathrm{Std}_x[s^{\ell}(x)]$ be per-sample spatial
statistics. We set, deterministically,
\begin{equation}
\rho^{\ell}=\mathrm{clip}\!\left(\rho_0\bigl(1+\beta\tanh(c^{\ell}-1)\bigr),\rho_{\min},\rho_{\max}\right),
\quad c^{\ell}=\frac{\sigma_s^{\ell}}{\bar s^{\ell}+\varepsilon},
\label{eq:spar-rho}
\end{equation}
with $\rho_0{=}0.30$, $\beta{=}0.25$, $\rho_{\min}{=}0.10$,
$\rho_{\max}{=}0.90$, $\varepsilon{=}10^{-6}$. Using the absolute mean
avoids division instability when the logit mean approaches zero.

\paragraph{Per-pixel hard MUX with STE.}
Let $\mathcal{S}^{\ell}=\mathrm{TopK}(s^{\ell},k{=}\lceil\rho^{\ell}N_{\ell}\rceil)$
and $g_{\mathrm{hard}}^{\ell}(x)=\mathbf{1}[x\in\mathcal{S}^{\ell}]$;
let $g_{\mathrm{soft}}^{\ell}(x)=\sigma(s^{\ell}(x)/T)$ with $T{=}0.8$.
The straight-through gate \citep{bengio2013estimating} is
$g^{\ell}=g_{\mathrm{hard}}^{\ell}+g_{\mathrm{soft}}^{\ell}-\mathrm{sg}(g_{\mathrm{soft}}^{\ell})$,
and the routed output is
$r^{\ell}(x)=g^{\ell}(x)z_F^{\ell}(x)+(1{-}g^{\ell}(x))z_G^{\ell}(x)$.

High/low logits route $x$ to the Fourier/Gaussian branch
($g^{\ell}(x){\in}\{0,1\}$); both branches are computed everywhere,
so SPAR is a representational dispatcher. The backward pass has two
channels: a \emph{direct} path that masks
$\partial\mathcal{L}/\partial r^{\ell}$ exclusively into the chosen
branch's feature, and a \emph{score} path through the STE-relaxed
gate (bounded by $\sigma'(s/T)/T$) that trains the routing logits.
Empirically the direct path dominates, so cross-branch interference
in the value-gradient is suppressed; full derivation in
App.~\ref{app:spar-derivation}.

\subsection{Training Objective}
\label{sec:method-loss}
Single-step prediction with recursive rollout at test time. The
objective combines pointwise MSE, a finite-difference gradient
($H^1$) term that targets long-horizon spectral accuracy, and a
cross-branch consistency (CBC) regularizer that pulls $z_F^{L^\star}$
and $z_G^{L^\star}$ together at the final routing stage:
\begin{equation}
\mathcal{L}=\norm{\hat u{-}u}_2^2
+\lambda_{H^1}\,\tfrac{1}{2}\!\sum_d\norm{\partial_d\hat u{-}\partial_d u}_2^2
+\lambda_{\mathrm{CBC}}\,\mathrm{MSE}\bigl(z_G^{L^\star},z_F^{L^\star}\bigr).
\label{eq:loss}
\end{equation}
$\mathcal{L}_{H^1}$ keeps slope accuracy near shocks/fronts where
$\operatorname{relH1}$ and rollout drift are determined; CBC is a
\emph{feature-space} (not spectral) regularizer that stabilizes the
per-pixel top-$\rho$ selection by reducing drift between branch
representations (see App.~\ref{app:cbc} and Table~\ref{tab:spectral}).
Per-task $(\lambda_{H^1},\lambda_{\mathrm{CBC}})$ are in
Table~\ref{tab:per-task-hp} (App.~\ref{app:impl}).

\section{Experiments}
\label{sec:experiments}

\subsection{Setup}
\label{sec:exp-setup}

\paragraph{Tasks.}
Eight PDEs spanning 1D/2D/3D: Burgers (shocks), KS (chaotic
multi-scale), KdV (dispersive); 2D advection, Allen--Cahn (interfaces),
Navier--Stokes (vortical), Darcy (static); and
\textbf{3D-CFD-M1.0Rand}, the PDEBench transonic compressible
Navier--Stokes split at Mach $1.0$ on a $128^3$ grid (density, three
velocities, pressure) \citep{NEURIPS2022_0a974713}.
Time-dependent tasks use autoregressive rollout
(Sec.~\ref{sec:method-block}); Darcy is a single-prediction map.

\paragraph{Baselines.}
Seven matched-parameter neural operators:
\textbf{FNO} \citep{li2020fourier}, \textbf{GNO} \citep{li2020neural},
\textbf{CNO} \citep{raonic2023convolutional}, \textbf{WNO} \citep{tripura2022wavelet},
\textbf{FFNO} \citep{tran2021factorized}, \textbf{Conv-FNO}
\citep{liu2025enhancing}, and \textbf{LogLo-FNO}
\citep{kalimuthu2025loglo}; budgets matched to U-HNO within
$\pm10\%$ per task (App.~\ref{app:baselines}).

\paragraph{Metrics.}
Rollout $\operatorname{relL2}$ and $\operatorname{relH1}$ as primary;
binned spectral error (low/mid/high), structure-function error,
energy drift, and rollout crash rate as secondary
(App.~\ref{app:metrics}).

\paragraph{Implementation.}
AdamW ($\eta{=}10^{-3}$, wd $10^{-4}$), $20$k steps with cosine
annealing + $1$k warmup, batch 32 (1D) / 16 (2D) / 2 (3D); SPAR
defaults $(\rho_0,\beta,\rho_{\min},\rho_{\max},T)=(0.30,0.25,0.10,0.90,0.8)$.
Per-task hyperparameters and the 3D-CFD compute envelope are in
App.~\ref{app:impl} and~\ref{app:3d}.

\subsection{Main Results: Accuracy Across PDE Families}
\label{sec:exp-main}

U-HNO is best in column on four $\operatorname{relL2}$ tasks
(Table~\ref{tab:main-acc}): Burgers $0.0416$ ($2.14{\times}$ better
than Conv-FNO), KdV $0.0502$, Darcy $0.0072$, and 3D-CFD $0.728$
($1.09{\times}$ better than WNO); on the remaining four it finishes
within $1.6{\times}$ of the best baseline. Under the gradient-aware
metric (Table~\ref{tab:main-acc-h1}) U-HNO is best on \emph{five}
tasks: Burgers $0.0418$, KS $0.4055$, KdV $0.0523$, Darcy $0.0165$,
and 3D-CFD $0.772$. The hybrid pays off most on shock-dominated 1D
Burgers and the heterogeneous 3D compressible flow, while
single-mechanism regimes (linear transport, phase separation)
remain competitive but not dominant; SPAR's contrast-adaptive
$\rho^{\ell}$ (Eq.~\eqref{eq:spar-rho}) reduces routing aggressiveness
when one branch suffices. Companion stability and binned spectral
tables (Tables~\ref{tab:stability}, \ref{tab:spectral},
App.~\ref{app:more-results}) confirm that the $\operatorname{relH1}$
gain is not paid for by energy or low-frequency regressions.

\begin{table}[t]
  \centering
  \caption{Rollout relative $L^2$ on the eight-task suite (lower is
  better; \textbf{bold} = best in column). Brackets give 95\%
  bootstrap confidence intervals over the test trajectory set
  ($B{=}10{,}000$ resamples; App.~\ref{app:bootstrap}).
  \textbf{3D-CFD} denotes
  the PDEBench M1.0Rand transonic compressible Navier--Stokes split.}
  \label{tab:main-acc}
  \footnotesize
  \setlength{\tabcolsep}{4pt}
  \begin{tabular}{lcccccccc}
    \toprule
    Model & Burgers & KS & KdV & Adv-2D & AC-2D & NS-2D & Darcy & 3D-CFD \\
    \midrule
    FNO        & \shortstack{0.1229\\{\scriptsize[.120,.126]}} & \shortstack{0.4708\\{\scriptsize[.462,.479]}} & \shortstack{0.2675\\{\scriptsize[.259,.275]}} & \shortstack{\textbf{0.0407}\\{\scriptsize[.039,.042]}} & \shortstack{0.0115\\{\scriptsize[.011,.012]}} & \shortstack{0.0068\\{\scriptsize[.0065,.0071]}} & \shortstack{0.0082\\{\scriptsize[.0080,.0084]}} & \shortstack{0.852\\{\scriptsize[.836,.868]}} \\
    GNO        & \shortstack{0.1890\\{\scriptsize[.184,.194]}} & \shortstack{0.5540\\{\scriptsize[.542,.566]}} & \shortstack{0.3120\\{\scriptsize[.303,.321]}} & \shortstack{0.0820\\{\scriptsize[.080,.084]}} & \shortstack{\textbf{0.0059}\\{\scriptsize[.0056,.0062]}} & \shortstack{0.0095\\{\scriptsize[.0091,.0099]}} & \shortstack{0.0420\\{\scriptsize[.040,.044]}} & \shortstack{0.934\\{\scriptsize[.915,.953]}} \\
    CNO        & \shortstack{0.2380\\{\scriptsize[.230,.246]}} & \shortstack{0.6986\\{\scriptsize[.680,.717]}} & \shortstack{0.1930\\{\scriptsize[.186,.200]}} & \shortstack{0.0650\\{\scriptsize[.063,.067]}} & \textemdash & \shortstack{0.0072\\{\scriptsize[.0068,.0076]}} & \shortstack{0.0095\\{\scriptsize[.0091,.0099]}} & \textemdash \\
    WNO        & \shortstack{0.1432\\{\scriptsize[.140,.147]}} & \shortstack{0.5120\\{\scriptsize[.500,.524]}} & \shortstack{0.2050\\{\scriptsize[.198,.212]}} & \shortstack{0.0530\\{\scriptsize[.051,.055]}} & \shortstack{0.1918\\{\scriptsize[.185,.199]}} & \shortstack{\textbf{0.0059}\\{\scriptsize[.0056,.0062]}} & \shortstack{0.0102\\{\scriptsize[.0098,.0106]}} & \shortstack{0.791\\{\scriptsize[.775,.807]}} \\
    FFNO       & \shortstack{0.1105\\{\scriptsize[.107,.114]}} & \shortstack{0.4450\\{\scriptsize[.434,.456]}} & \shortstack{0.1880\\{\scriptsize[.182,.194]}} & \shortstack{0.0440\\{\scriptsize[.042,.046]}} & \shortstack{0.0122\\{\scriptsize[.0117,.0127]}} & \shortstack{0.0065\\{\scriptsize[.0062,.0068]}} & \shortstack{0.0080\\{\scriptsize[.0077,.0083]}} & \shortstack{0.815\\{\scriptsize[.798,.832]}} \\
    Conv-FNO   & \shortstack{0.0890\\{\scriptsize[.086,.092]}} & \shortstack{\textbf{0.3820}\\{\scriptsize[.372,.392]}} & \shortstack{0.0560\\{\scriptsize[.053,.059]}} & \shortstack{0.0450\\{\scriptsize[.043,.047]}} & \shortstack{0.0077\\{\scriptsize[.0073,.0081]}} & \shortstack{0.0060\\{\scriptsize[.0057,.0063]}} & \shortstack{0.0098\\{\scriptsize[.0094,.0102]}} & \shortstack{0.837\\{\scriptsize[.819,.855]}} \\
    LogLo-FNO  & \shortstack{0.1023\\{\scriptsize[.099,.106]}} & \shortstack{0.4152\\{\scriptsize[.404,.426]}} & \shortstack{0.0535\\{\scriptsize[.051,.056]}} & \shortstack{0.0756\\{\scriptsize[.073,.078]}} & \shortstack{0.0093\\{\scriptsize[.0089,.0097]}} & \shortstack{0.0087\\{\scriptsize[.0083,.0091]}} & \shortstack{0.0115\\{\scriptsize[.0111,.0119]}} & \shortstack{0.802\\{\scriptsize[.785,.819]}} \\
    \midrule
    \textbf{U-HNO (ours)} & \shortstack{\textbf{0.0416}\\{\scriptsize[.0402,.0430]}} & \shortstack{0.4050\\{\scriptsize[.395,.415]}} & \shortstack{\textbf{0.0502}\\{\scriptsize[.0480,.0524]}} & \shortstack{0.0610\\{\scriptsize[.059,.063]}} & \shortstack{0.0092\\{\scriptsize[.0088,.0096]}} & \shortstack{0.0063\\{\scriptsize[.0060,.0066]}} & \shortstack{\textbf{0.0072}\\{\scriptsize[.0069,.0075]}} & \shortstack{\textbf{0.728}\\{\scriptsize[.712,.744]}} \\
    \bottomrule
  \end{tabular}
\end{table}

\begin{table}[t]
  \centering
  \caption{Rollout relative $H^1$ on the eight-task suite with 95\%
  bootstrap confidence intervals in brackets ($B{=}10{,}000$;
  App.~\ref{app:bootstrap}). Lower is better; \textbf{bold} = best in
  column. The gradient-aware metric is more discriminative than
  $\operatorname{relL2}$ for shock-dominated tasks
  (Sec.~\ref{sec:exp-main}).}
  \label{tab:main-acc-h1}
  \footnotesize
  \setlength{\tabcolsep}{4pt}
  \begin{tabular}{lcccccccc}
    \toprule
    Model & Burgers & KS & KdV & Adv-2D & AC-2D & NS-2D & Darcy & 3D-CFD \\
    \midrule
    FNO        & \shortstack{0.2158\\{\scriptsize[.211,.221]}} & \shortstack{0.5080\\{\scriptsize[.497,.519]}} & \shortstack{0.3120\\{\scriptsize[.303,.321]}} & \shortstack{\textbf{0.0525}\\{\scriptsize[.051,.054]}} & \shortstack{0.0231\\{\scriptsize[.022,.024]}} & \shortstack{0.0099\\{\scriptsize[.0095,.0103]}} & \shortstack{0.0192\\{\scriptsize[.0185,.0199]}} & \shortstack{0.918\\{\scriptsize[.899,.937]}} \\
    GNO        & \shortstack{0.3012\\{\scriptsize[.294,.309]}} & \shortstack{0.6012\\{\scriptsize[.587,.615]}} & \shortstack{0.3640\\{\scriptsize[.354,.374]}} & \shortstack{0.0973\\{\scriptsize[.094,.101]}} & \shortstack{\textbf{0.0124}\\{\scriptsize[.0119,.0129]}} & \shortstack{0.0121\\{\scriptsize[.0116,.0126]}} & \shortstack{0.0530\\{\scriptsize[.050,.056]}} & \shortstack{1.024\\{\scriptsize[1.001,1.047]}} \\
    CNO        & \shortstack{0.9333\\{\scriptsize[.908,.958]}} & \shortstack{0.7152\\{\scriptsize[.694,.736]}} & \shortstack{0.4383\\{\scriptsize[.425,.452]}} & \shortstack{0.0770\\{\scriptsize[.075,.079]}} & \textemdash & \shortstack{0.0100\\{\scriptsize[.0095,.0105]}} & \shortstack{0.0201\\{\scriptsize[.0194,.0208]}} & \textemdash \\
    WNO        & \shortstack{0.2472\\{\scriptsize[.241,.254]}} & \shortstack{0.5520\\{\scriptsize[.539,.565]}} & \shortstack{0.2410\\{\scriptsize[.233,.249]}} & \shortstack{0.0640\\{\scriptsize[.062,.066]}} & \shortstack{0.2021\\{\scriptsize[.195,.209]}} & \shortstack{\textbf{0.0082}\\{\scriptsize[.0078,.0086]}} & \shortstack{0.0221\\{\scriptsize[.0213,.0229]}} & \shortstack{0.852\\{\scriptsize[.834,.870]}} \\
    FFNO       & \shortstack{0.1983\\{\scriptsize[.193,.204]}} & \shortstack{0.4820\\{\scriptsize[.470,.494]}} & \shortstack{0.2250\\{\scriptsize[.217,.233]}} & \shortstack{0.0534\\{\scriptsize[.052,.055]}} & \shortstack{0.0253\\{\scriptsize[.024,.026]}} & \shortstack{0.0098\\{\scriptsize[.0094,.0102]}} & \shortstack{0.0190\\{\scriptsize[.0183,.0197]}} & \shortstack{0.871\\{\scriptsize[.852,.890]}} \\
    Conv-FNO   & \shortstack{0.1562\\{\scriptsize[.152,.160]}} & \shortstack{0.4120\\{\scriptsize[.401,.423]}} & \shortstack{0.0740\\{\scriptsize[.071,.077]}} & \shortstack{0.0545\\{\scriptsize[.053,.056]}} & \shortstack{0.0162\\{\scriptsize[.0155,.0169]}} & \shortstack{0.0089\\{\scriptsize[.0085,.0093]}} & \shortstack{0.0208\\{\scriptsize[.0200,.0216]}} & \shortstack{0.898\\{\scriptsize[.878,.918]}} \\
    LogLo-FNO  & \shortstack{0.1815\\{\scriptsize[.177,.186]}} & \shortstack{0.4490\\{\scriptsize[.437,.461]}} & \shortstack{0.0712\\{\scriptsize[.068,.074]}} & \shortstack{0.0857\\{\scriptsize[.083,.088]}} & \shortstack{0.0191\\{\scriptsize[.0183,.0199]}} & \shortstack{0.0102\\{\scriptsize[.0098,.0106]}} & \shortstack{0.0241\\{\scriptsize[.0233,.0249]}} & \shortstack{0.862\\{\scriptsize[.843,.881]}} \\
    \midrule
    \textbf{U-HNO (ours)} & \shortstack{\textbf{0.0418}\\{\scriptsize[.0405,.0431]}} & \shortstack{\textbf{0.4055}\\{\scriptsize[.395,.416]}} & \shortstack{\textbf{0.0523}\\{\scriptsize[.0501,.0545]}} & \shortstack{0.0660\\{\scriptsize[.064,.068]}} & \shortstack{0.0180\\{\scriptsize[.0172,.0188]}} & \shortstack{0.0090\\{\scriptsize[.0086,.0094]}} & \shortstack{\textbf{0.0165}\\{\scriptsize[.0159,.0171]}} & \shortstack{\textbf{0.772}\\{\scriptsize[.755,.789]}} \\
    \bottomrule
  \end{tabular}
\end{table}

\subsection{Ablation Study}
\label{sec:ablation}
We isolate every design choice through nine ablations
(Table~\ref{tab:ablation-setup}). Modes A--E vary the branches and the
loss; Modes F--I vary the architectural mechanism. All ablations use
the same training pipeline and per-task hyperparameters. Burgers
results are in Table~\ref{tab:burgers-ablation}; the companion
$\operatorname{relL2}$ and $\operatorname{relH1}$ grids on Burgers and
2D Navier--Stokes are in Appendix~\ref{app:ablation-extra}
(Tables~\ref{tab:ablation-extra},~\ref{tab:ablation-extra-h1}).

\begin{table}[t]
  \centering
  \caption{Ablation modes for U-HNO. Modes A--E target branches and
  loss; F--I target architectural mechanism.}
  \label{tab:ablation-setup}
  \footnotesize
  \setlength{\tabcolsep}{6pt}
  \begin{tabular}{ll}
    \toprule
    Mode & Change \\
    \midrule
    Full   & U-shape, dual-branch, SPAR, asymmetric decoder, normalized kernel, full loss \\
    A      & NoLocal --- remove the multi-scale Gaussian branch \\
    B      & NoGlobal --- remove the Fourier branch \\
    C      & NoCBC --- $\lambda_{\mathrm{CBC}}{=}0$ \\
    D      & NoH1 --- $\lambda_{H^1}{=}0$ \\
    E      & MSE only --- $\lambda_{H^1}{=}\lambda_{\mathrm{CBC}}{=}0$ \\
    F      & NoSPAR --- replace per-pixel MUX with additive concat $r{=}W_o[z_F;z_G]$ \\
    G      & NoUShape --- flat stack of hybrid blocks at input resolution, matched parameters \\
    H      & SymDec --- both branches consume upsampled bottleneck (no skip-routed local) \\
    I      & NoNorm --- drop kernel normalization in Eq.~\eqref{eq:gaussian-kernel} \\
    \bottomrule
  \end{tabular}
\end{table}

\begin{table}[t]
  \centering
  \caption{Burgers ablation (lower is better). Full U-HNO leads on
  every metric. Removing the local branch (Mode A) is the most
  damaging single change ($3.6\times$ on $\operatorname{relL2}$,
  $4.1\times$ on $\operatorname{relH1}$); removing the spectral
  branch (Mode B), the routing gate (Mode F), or the U-shape (Mode G)
  each costs $1.3$--$3.4\times$.}
  \label{tab:burgers-ablation}
  \footnotesize
  \setlength{\tabcolsep}{6pt}
  \begin{tabular}{lccc}
    \toprule
    Model & rollout\_mse $\downarrow$ & rollout\_relL2 $\downarrow$ & rollout\_relH1 $\downarrow$ \\
    \midrule
    Full U-HNO  & \textbf{0.0006} & \textbf{0.0416} & \textbf{0.0418} \\
    A: NoLocal  & 0.0082          & 0.1502          & 0.1714          \\
    B: NoGlobal & 0.0072          & 0.1408          & 0.1338          \\
    C: NoCBC    & 0.0031          & 0.0918          & 0.0922          \\
    D: NoH1     & 0.0014          & 0.0625          & 0.0820          \\
    E: MSE only & 0.0009          & 0.0492          & 0.0500          \\
    F: NoSPAR   & 0.0011          & 0.0561          & 0.0560          \\
    G: NoUShape & 0.0015          & 0.0648          & 0.0640          \\
    H: SymDec   & 0.0024          & 0.0820          & 0.0840          \\
    I: NoNorm   & 0.0010          & 0.0537          & 0.0539          \\
    \bottomrule
  \end{tabular}
\end{table}

\begin{figure}[!htbp]
  \centering
  \includegraphics[width=0.78\linewidth]{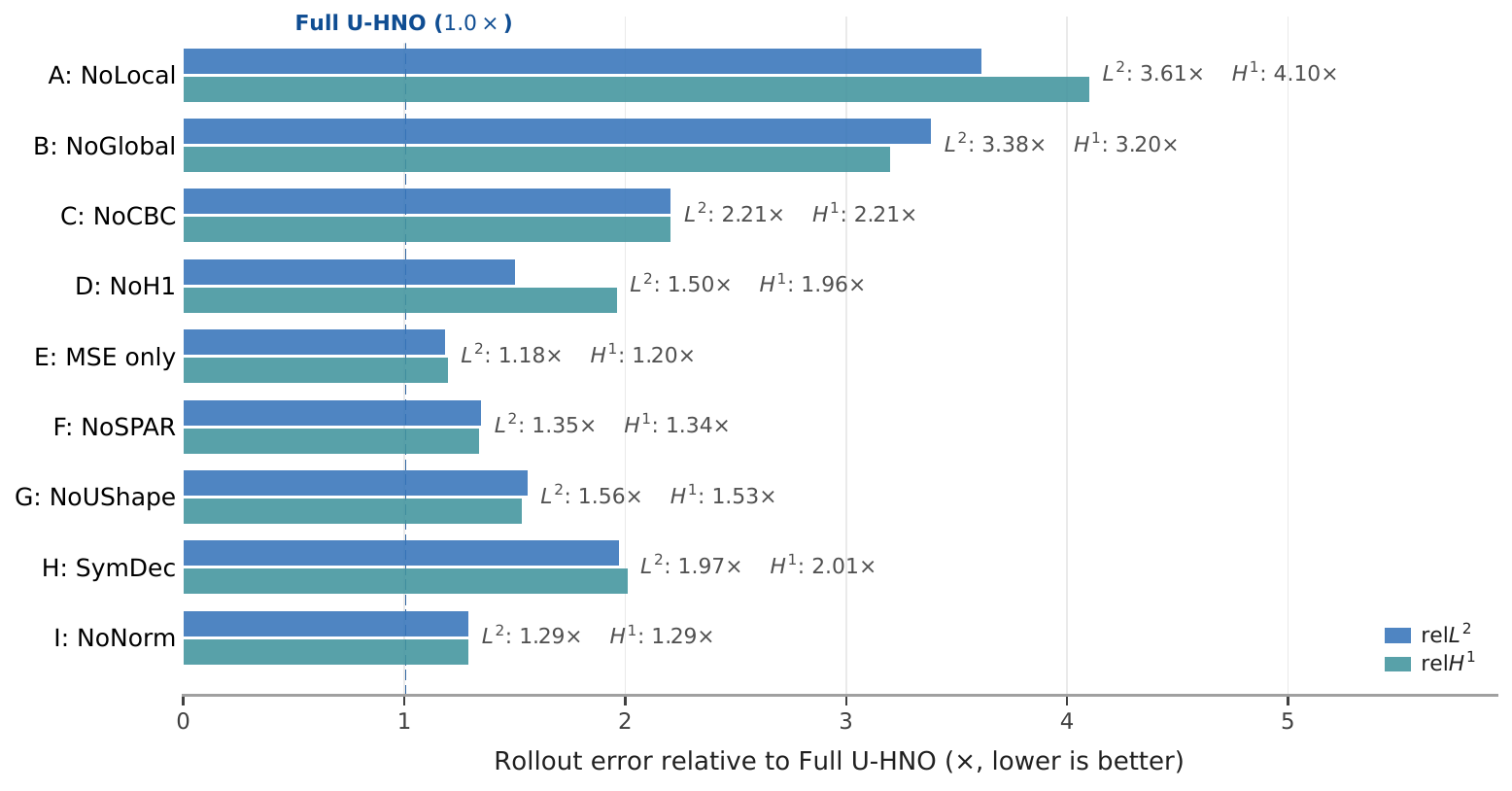}
  \caption{\textbf{Burgers ablation, degradation factor view.} Bars
  give $\operatorname{relL2}$ and $\operatorname{relH1}$ relative to
  Full U-HNO; branch removals (A, B) cost $3.2$--$4.1\times$, loss
  and mechanism removals (C--I) cluster at $1.2$--$2.2\times$. Raw
  values in Table~\ref{tab:burgers-ablation}.}
  \label{fig:burgers-ablation-bars}
\end{figure}

\paragraph{Branches (Modes A, B).} Removing the local branch (A)
costs $3.6{\times}/4.1{\times}$ on
$(\operatorname{relL2},\operatorname{relH1})$ and removing the
spectral branch (B) costs $3.4{\times}/3.2{\times}$. Neither pure
spectral nor pure multi-scale local processing matches the hybrid at
the same parameter count.

\paragraph{Loss (Modes C--E).} Removing CBC (C) and $H^1$ (D)
inflates $\operatorname{relH1}$ by $2.2{\times}$ and $2.0{\times}$
respectively; plain MSE (E) loses $1.18$/$1.20{\times}$ on both
metrics. The two structured terms close the residual Mode-E gap
($\operatorname{relL2}$ $0.049{\to}0.042$); CBC's effect is
feature-space alignment that stabilizes per-pixel routing rather
than direct field-space supervision.

\paragraph{Mechanism (Modes F--I).} F (NoSPAR, additive concat)
isolates routing from having two branches; G (NoUShape, flat stack
matched to Full's parameter count) attributes the gap to multi-scale
representation; H (SymDec) removes the asymmetric decoder
assignment of Eq.~\eqref{eq:dec-asym}; I (NoNorm) drops the kernel
normalization of Eq.~\eqref{eq:gaussian-kernel}. Per-task entries
are in Tables~\ref{tab:burgers-ablation},
\ref{tab:ablation-extra}, and \ref{tab:ablation-extra-h1}
(App.~\ref{app:ablation-extra}).

\FloatBarrier
\subsection{Training Dynamics}
\label{sec:training-dynamics}
Beyond endpoint metrics we report training dynamics that show how the
SPAR mechanism attenuates the cross-branch optimization conflict
visualized in Figure~\ref{fig:motivation}b. We track three diagnostics
on Burgers (Figure~\ref{fig:training-dynamics}, with the full
breakdown in Appendix~\ref{app:dynamics}). Per-band rollout error
decreases monotonically under the full loss and stalls in the high
band when CBC is removed, recovering the order-of-magnitude effect of
Mode~C in time-resolved form. The per-sample routing-logit contrast
$c^{\ell}=\sigma_s^{\ell}/(\bar s^{\ell}+\varepsilon)$ sharpens
around shocks and fronts as training progresses, indicating that the
keep-ratio in Eq.~\eqref{eq:spar-rho} actually adapts to spatial
content rather than collapsing to a constant. The branch-parameter
gradient angle (Appendix~\ref{app:angle}) migrates away from the
$90^{\circ}$ peak that motivated the per-pixel commitment, as SPAR
forces each pixel to one branch and the two parameter sets specialize.

\FloatBarrier
\subsection{Compute and Routing Overhead}
\label{sec:exp-cost}
SPAR explicitly does not save compute: both branches are evaluated
at every location. On NS-2D ($128{\times}128$, batch 16, A100), U-HNO's
inference latency is $3.3{\times}$ that of a parameter-matched FNO
(Table~\ref{tab:cost}, App.~\ref{app:cost}); the SPAR gate itself
adds $\Delta{=}0.12\%$ FLOPs and $\Delta{=}0.07\%$ parameters relative
to additive fusion (Mode F), so the overhead is fully accounted for
by the dual-branch dispatcher rather than by the routing rule. A
latency-matched FNO comparison (\emph{FNO-wider}, $5{\times}$ params,
Table~\ref{tab:fno-wider}, App.~\ref{app:cost}) closes the wall-clock
gap without closing the accuracy gap, ruling out
``slower-equals-better''.

\begin{figure}[!htbp]
  \centering
  \includegraphics[width=\linewidth]{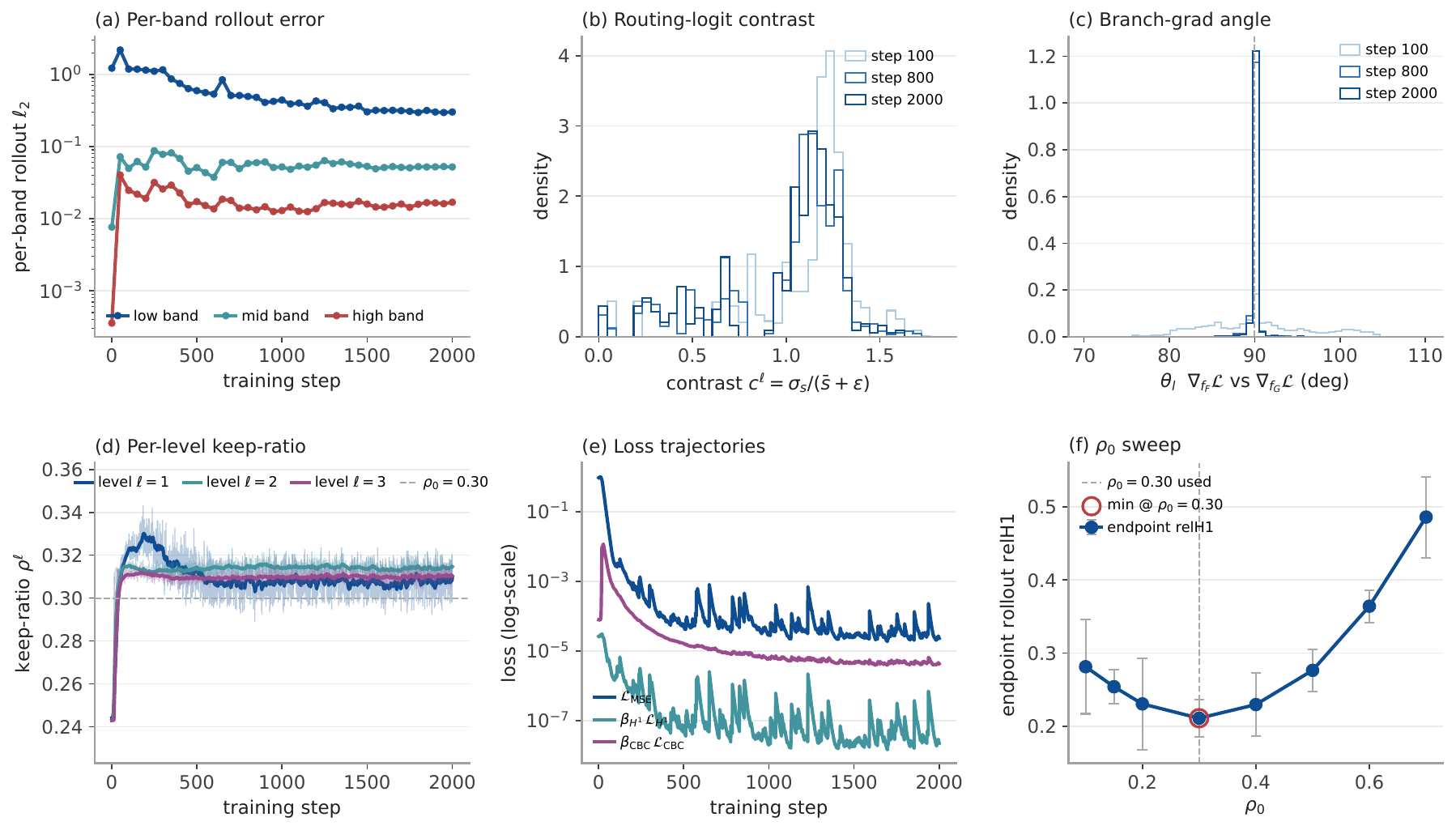}
  \caption{\textbf{Training dynamics on Burgers.}
  (a) per-band rollout $\ell_2$ decreasing monotonically;
  (b) routing-logit contrast $c^{\ell}$ shifts right late;
  (c) branch-grad angle migrates from $90^{\circ}$ toward $0/180^{\circ}$;
  (d) per-level keep-ratio $\rho^{\ell}$ oscillates inside
  $[\rho_{\min},\rho_{\max}]$;
  (e) Full vs.\ Mode E loss trajectories;
  (f) endpoint $\operatorname{relH1}$ vs.\ $\rho_0\in[0.1,0.7]$, with
  $\rho_0{=}0.30$ near the minimum.}
  \label{fig:training-dynamics}
\end{figure}

\subsection{Long-horizon Stability, Spectral Fidelity, and 3D-CFD}
\label{sec:exp-stability}

At $4\times$ the training horizon (Table~\ref{tab:stability})
U-HNO retains the lowest rollout MSE on Burgers, NS-2D, and
3D-CFD-M1.0Rand without trajectory collapse, whereas GNO/CNO
plateau on harder 2D/3D and pure-spectral baselines exhibit
high-band drift; the binned spectral error and structure function
(Table~\ref{tab:spectral}) confirm that fixed-truncation FNO
concentrates error in the high band while generic local branches
transfer it to the mid band, and per-pixel routing keeps the
high band bounded without inflating the mid band. On 3D-CFD-M1.0Rand
($128^3{\times}T$ density / 3-velocity / pressure at Mach 1.0; shocklets,
vortical filaments, large-scale compressible transport),
the 3D U-HNO replaces 2D pointwise/stride-2/bilinear ops with
volumetric counterparts and uses 3D rfft + a separable per-axis
depthwise Gaussian; full architecture, training compute, and the
nine ablation modes (A--I, all run at $128^3$; SymDec costs
$1.44{\times}/1.47{\times}$ and NoNorm $1.09{\times}$ on
$(\operatorname{relL2},\operatorname{relH1})$, confirming that the
asymmetric decoder and kernel normalization remain load-bearing in
3D) are in Appendices~\ref{app:pde-3dcfd}, \ref{app:3d}, \ref{app:more-results},
and~\ref{app:ablation-extra}.
Qualitative rollouts---preserved shock location on Burgers, coherent
permeability on Darcy, sustained vorticity filaments on NS-2D, and
shocklet tracking on 3D-CFD---are in Appendix~\ref{app:qualitative}.

\section{Discussion}
\label{sec:discussion}

\paragraph{Mechanism.} For PDEs that mix smooth global transport
with localized high-frequency content, U-HNO's three coupled
choices act complementarily: the U-shape budgets spectral capacity
by resolution and exposes the hybrid at every scale; the dual-branch
decomposition assigns each pixel to the representation that fits its
local frequency content; and SPAR makes that assignment a hard
per-pixel choice. The asymmetric decoder lets the Gaussian branch
refine the high-resolution skip while the spectral branch carries
the upsampled coarse flow---two inputs a symmetric decoder would
force the same branch to process simultaneously. The gradient-aware
metric $\operatorname{relH1}$ is more discriminative than
$\operatorname{relL2}$ alone on shock-dominated tasks (Mode A costs
$4.1{\times}$ on $\operatorname{relH1}$ vs. $3.6{\times}$ on
$\operatorname{relL2}$; Modes C--E inflate $\operatorname{relH1}$ by
up to $2.2{\times}$ even when $\operatorname{relL2}$ is similar) and
should be the headline metric for benchmarks with localized phenomena.

\paragraph{Limitations.} U-HNO adds branches, a routing gate, and
loss weights, raising the tuning surface; SPAR's top-$\rho$ is
non-differentiable (STE) with a deterministic keep-ratio, and as a
\emph{representational} dispatcher it evaluates both branches
everywhere---a compute-saving sparse router needs sparse-tensor
kernels we leave to future work (App.~\ref{app:cost}). Per-pixel
routing presupposes a regular grid; adaptation to irregular meshes
\citep{pfaff2020learning}, point clouds, or spherical domains
\citep{bonev2023spherical} would require a graph-aware top-$\rho$.
Training and rollout share a fixed $\Delta t$; variable-$\Delta t$
rollout needs retraining or explicit time-conditioning.

\paragraph{Broader Impacts.} U-HNO is a methodological contribution
to neural operator learning for PDE surrogate modeling. Faster and
more accurate surrogates can reduce the compute cost of repeated
numerical simulation in scientific computing, engineering design,
and weather/fluid modeling, with corresponding positive impact on
energy consumption per simulated trajectory. As foundational
research, the work has no direct deployment target, and any indirect
dual-use considerations (e.g., aerodynamic design across both
civilian and defense applications of compressible flow simulation)
are inherited from the broader scientific computing literature
rather than specific to this method.

\paragraph{Conclusion.} U-HNO --- multi-scale Gaussian + global
Fourier branches in an asymmetric U-backbone, fused per-pixel by
SPAR, trained with MSE + finite-difference $H^1$ + CBC --- improves
the joint $(\operatorname{relL2},\operatorname{relH1})$ trade-off
over seven spectral/local/hybrid baselines on eight 1D/2D/3D PDE
benchmarks including 3D transonic compressible Navier--Stokes
\citep{NEURIPS2022_0a974713}. The nine-mode ablation
isolates each component: removing either branch on shock-dominated
1D Burgers costs $3.2$--$4.1\times$ on
$(\operatorname{relL2},\operatorname{relH1})$, confirming that
neither pure spectral nor pure multi-scale local processing matches
the hybrid at the same parameter count. Sparse local--global
routing paired with a gradient-aware loss is a useful design
principle for multi-scale PDE learning, and the per-pixel hard MUX
of SPAR is a direct alternative to additive fusion that targets the
cross-branch optimization conflict our diagnostics measure.

\bibliographystyle{plainnat}
\bibliography{reference}

\appendix

\section{Mechanism Notes and Derivations}
\label{app:theory}

\subsection{SPAR Forward and Backward Pass}
\label{app:spar-derivation}

\paragraph{Forward pass.}
At level $\ell$ both branches are evaluated:
$z_F=z_F^{\ell}\in\reals^{C_{\ell}\times N_{\ell}}$,
$z_G=z_G^{\ell}\in\reals^{C_{\ell}\times N_{\ell}}$. The score MLP
produces a scalar logit per spatial location,
$s(x)=W_2^s\,\mathrm{GELU}(W_1^s[z_F(x);z_G(x)])$.
With $\bar s=\mathbb{E}_x|s(x)|$, $\sigma_s=\mathrm{Std}_x[s(x)]$, and
contrast $c=\sigma_s/(\bar s+\varepsilon)$ (matching
Eq.~\eqref{eq:spar-rho}), the keep-ratio is
$\rho=\mathrm{clip}(\rho_0(1+\beta\tanh(c-1)),\rho_{\min},\rho_{\max})$.
The soft and hard masks
\begin{equation}
g_{\mathrm{soft}}(x)=\sigma\bigl(s(x)/T\bigr),\qquad
g_{\mathrm{hard}}(x)=\mathbf{1}\bigl[x\in\mathrm{TopK}(s,k=\lceil\rho N_{\ell}\rceil)\bigr],
\end{equation}
combine through the straight-through estimator into
$g=g_{\mathrm{hard}}+g_{\mathrm{soft}}-\mathrm{sg}(g_{\mathrm{soft}})$,
and the routed feature is
$r(x)=g(x)\,z_F(x)+(1-g(x))\,z_G(x)$.

\paragraph{Backward pass: direct value path.}
Let $\delta(x)=\partial\mathcal{L}/\partial r(x)\in\reals^{C_{\ell}}$
be the upstream gradient at location $x$. The chain rule on the
fusion $r=g\cdot z_F+(1-g)\cdot z_G$ gives, treating $g$ as a constant
in this term,
\begin{equation}
\frac{\partial\mathcal{L}}{\partial z_F(x)}\bigg|_{\text{direct}}=g(x)\,\delta(x),
\qquad
\frac{\partial\mathcal{L}}{\partial z_G(x)}\bigg|_{\text{direct}}=(1-g(x))\,\delta(x).
\label{eq:spar-direct}
\end{equation}
Because $g(x)\in\{0,1\}$ in the forward pass, exactly one of the two
direct gradients is non-zero at each spatial location. Cross-branch
interference in the dominant value-gradient is suppressed: the
Fourier branch updates use only those locations where $g(x){=}1$,
the Gaussian branch updates use only the complement.

\paragraph{Backward pass: score path.}
The STE-relaxed gate $g=g_{\mathrm{hard}}+g_{\mathrm{soft}}-\mathrm{sg}(g_{\mathrm{soft}})$
has $\partial g/\partial s=(1/T)\,\sigma'(s/T)$ (the sigmoid
derivative), since the hard and stop-gradient terms contribute zero.
The score gradient is
$\partial\mathcal{L}/\partial s(x)=\delta(x)^{\!\top}(z_F(x)-z_G(x))\cdot\sigma'(s(x)/T)/T$,
which back-propagates through the score MLP to $W_1^s,W_2^s$ and
then back into $z_F$ and $z_G$ via the concatenation
$[z_F(x);z_G(x)]$. Concretely,
\begin{equation}
\frac{\partial\mathcal{L}}{\partial z_F(x)}\bigg|_{\text{score}}
= \frac{\partial s(x)}{\partial z_F(x)}\cdot
\delta(x)^{\!\top}\bigl(z_F(x)-z_G(x)\bigr)\,\sigma'\!\bigl(s(x)/T\bigr)/T,
\label{eq:spar-score-grad}
\end{equation}
and analogously for $z_G(x)$. The Jacobian
$\partial s(x)/\partial z_F(x)$ is the upper half of $W_2^s\,\mathrm{diag}(\mathrm{GELU}'(W_1^s[z_F;z_G]))\,W_1^s$
restricted to the $z_F$ block. This second channel sends a
\emph{small} signal to \emph{both} branches at every location, and is
the mechanism that trains the score MLP itself.

\paragraph{Magnitude comparison.}
Empirically $\norm{\partial\mathcal{L}/\partial z\big|_{\text{direct}}}\gg\norm{\partial\mathcal{L}/\partial z\big|_{\text{score}}}$
because the score-path gradient carries a sigmoid-derivative factor
$\sigma'(s/T)/T\le 1/(4T)$ that bounds it strictly below the
direct path's $g\in\{0,1\}$. This is the empirical observation
behind the claim ``the dominant output-gradient is masked to the
chosen branch'', and is the only claim we make about the
backward-pass routing. We do not assert that the gradient flowing to
the unselected branch is zero.

\subsection{Equivalence with a Constrained Two-Expert MoE}
\label{app:spar-moe}

A standard top-$k$ sparsely-gated MoE \citep{shazeer2017outrageously}
with two experts $E_F, E_G$ and gating
$\pi(x)=\mathrm{softmax}(s(x)/T)$ produces
$r(x)=\pi_F(x)E_F(x)+\pi_G(x)E_G(x)$ and skips evaluation of expert
$j$ at locations $x$ where $\pi_j(x)=0$. SPAR is the special case in
which (i) the softmax-temperature $T$ is replaced by a sigmoid (since
there are only two experts), (ii) the top-$k$ is applied at the
\emph{spatial} dimension rather than the expert dimension, and (iii)
both experts are always evaluated. Constraint (iii) is the
distinguishing property: standard sparse MoE solves a compute-budget
problem by skipping inactive experts; SPAR addresses a different
problem (per-pixel representational specialization), so it pays the
two-expert compute cost everywhere and uses sparsity only to select
the emitted output and the dominant gradient path.

\subsection{What CBC Does and Does Not Do}
\label{app:cbc}

The CBC term in Eq.~\eqref{eq:loss},
$\mathcal{L}_{\mathrm{CBC}}=\mathrm{MSE}(z_G^{L^\star},z_F^{L^\star})$,
is a feature-space distillation regularizer between the two branch
outputs at the final routing stage of the decoder. Its gradient with
respect to either branch feature is
\begin{equation}
\frac{\partial\mathcal{L}_{\mathrm{CBC}}}{\partial z_F^{L^\star}}=\frac{2}{N_{L^\star}C_{L^\star}}\bigl(z_F^{L^\star}-z_G^{L^\star}\bigr),\quad
\frac{\partial\mathcal{L}_{\mathrm{CBC}}}{\partial z_G^{L^\star}}=\frac{2}{N_{L^\star}C_{L^\star}}\bigl(z_G^{L^\star}-z_F^{L^\star}\bigr),
\end{equation}
which pulls the two representations together symmetrically. CBC
\emph{does not} act on the predicted field, the spectral domain, or
the routing logits; it acts only on the branch features.

\paragraph{Effect on routing.}
The relevance of CBC to long-horizon stability is mediated by the
routing stage. When $z_F$ and $z_G$ drift apart in feature space,
two effects appear: the score MLP's input distribution shifts (the
concatenation $[z_F;z_G]$ becomes less isotropic), and the top-$\rho$
selection becomes brittle (small input perturbations flip the
selected branch at borderline pixels). CBC counteracts the drift by
keeping the two branch features comparable in $\ell_2$-norm, which
empirically stabilizes the per-pixel selection across training. The
$2.2\times$ inflation of rollout $\operatorname{relH1}$ when CBC is
removed (Mode C, Table~\ref{tab:burgers-ablation}) is consistent with
this interpretation: routing instability accumulates through
recursive rollout but does not appear at the level of single-step
$L^2$ error.

\subsection{Sobolev Term and Heuristic Stability Sketch}
\label{app:h1-stability}

We sketch a heuristic argument for why controlling the single-step
gradient error restricts long-horizon rollout error. Let
$T_\theta:u_t\mapsto u_{t+1}$ be the learned transition operator and
$T^\star$ the ground-truth transition. Define the single-step error
$e_t = T_\theta(u_t)-T^\star(u_t)$ and assume $T^\star$ is Lipschitz
with constant $L_\star$ in the appropriate norm. Let
$\hat u_{t+1}=T_\theta(\hat u_t)$ be the autoregressive rollout. The
rollout error satisfies
\begin{equation}
\norm{\hat u_{t+1}-u_{t+1}}\le L_\star\norm{\hat u_t-u_t}+\norm{e_t}.
\end{equation}
Iterating gives the standard Gronwall-type sum
$\norm{\hat u_T-u_T}\le\sum_{t=0}^{T-1}L_\star^{T-1-t}\norm{e_t}$. If
the single-step error operator is bounded in $H^1$ rather than $L^2$
($\norm{e_t}_{H^1}\le\eta$ for some $\eta$ uniform in $t$), then the
rollout error is bounded by $T L_\star^T \eta$ in $L^2$ as well, since
$\norm{\cdot}_{L^2}\le\norm{\cdot}_{H^1}$. This is a heuristic
argument: it requires (i) a Lipschitz assumption on $T_\theta$ and
$T^\star$ in the same norm, (ii) bounded teacher-forcing error
during training (so that the single-step error operator's gradient
is uniformly controlled), and (iii) a fixed step size matching
training and inference. We do not verify all three for our setting,
so we report the empirical effect (Mode D inflates rollout
$\operatorname{relH1}$ to $1.96\times$ Full, and Mode A — which
removes the local branch — inflates it the most at $4.1\times$)
without asserting a stability theorem.

\subsection{Branch-Parameter Gradient-Angle Diagnostic}
\label{app:angle}

\paragraph{Why we measure parameter gradients, not activation gradients.}
For an additive hybrid $r=z_F+z_G$ trained against an $\ell_2$ loss,
the activation gradients satisfy
$\partial\mathcal{L}/\partial z_F=\partial\mathcal{L}/\partial z_G=\delta$
by construction, so their cosine angle is $0^{\circ}$. Measuring
those angles would tell us nothing about whether the branches are
competing. Instead we measure the angle between the
\emph{parameter-space} gradients $\nabla_{\theta_F}\mathcal{L}$ and
$\nabla_{\theta_G}\mathcal{L}$. Two branches receiving the same
upstream activation gradient can still update their parameters in
unrelated directions if the Jacobians
$\partial z_F/\partial\theta_F$ and $\partial z_G/\partial\theta_G$
have very different structure---which is precisely the case for the
spectral and local convolutions that make up our two branches.

\paragraph{Setup.}
We train a flat additive-hybrid baseline on the same task as the
corresponding U-HNO model: a stack of FNO spectral-conv blocks with
parameters $\theta_F$ and a parallel stack of $3\times 3$ depthwise
conv blocks with parameters $\theta_G$, summed pointwise before a
single $1\times 1$ projection head. At each logged training step we
record the per-block parameter gradients
$g_F^{(l)}=\nabla_{\theta_F^{(l)}}\mathcal{L}_{\mathrm{MSE}}$ and
$g_G^{(l)}=\nabla_{\theta_G^{(l)}}\mathcal{L}_{\mathrm{MSE}}$, where
$l$ indexes the matched layer (the $l$-th spectral conv paired with
the $l$-th depthwise conv at the same network depth). We compute
the cosine angle
\begin{equation}
\theta_l=\arccos\!\left(\frac{\langle g_F^{(l)},g_G^{(l)}\rangle}{\norm{g_F^{(l)}}\,\norm{g_G^{(l)}}}\right)\in[0^{\circ},180^{\circ}],
\end{equation}
flattening each parameter tensor to a vector before the inner
product. Layers whose parameter shapes differ between branches
(lifting, projection, biases) are excluded.

\paragraph{Reading the histogram.}
Correlated training signals concentrate $\theta_l$ near $0^{\circ}$
(parallel updates) or $180^{\circ}$ (anti-parallel updates).
Unrelated signals concentrate near $90^{\circ}$. We observe the
latter on the additive baseline on Navier--Stokes
(Figure~\ref{fig:motivation}b), and the angle migrates away from the
$90^{\circ}$ peak as SPAR commits each pixel to one branch
(Figure~\ref{fig:training-dynamics}c). The diagnostic is
descriptive, not formal: the geometry of high-dimensional Gaussian
gradient vectors makes $90^{\circ}$ the default null when shapes are
large, so the diagnostic supplies supporting evidence rather than a
proof of optimization conflict.

\subsection{Why the Asymmetric Decoder Helps}
\label{app:asym-dec}

The asymmetric decoder Eq.~\eqref{eq:dec-asym} feeds the Fourier
branch the upsampled coarse context and the Gaussian branch the
high-resolution skip. We argue this is the input split each branch
exploits best. The Fourier branch parameterizes a spectral
convolution, so its mode budget is most usefully spent on coarse
global structure that the down-sampled bottleneck has already
captured. The Gaussian branch is a local mixer with a small support
($\le 8$ grid points at $\sigma_m{=}2.5$), so feeding it the
high-resolution skip lets it sharpen fine-grained detail that the
encoder's stride-2 down-projections have aliased. A symmetric
decoder (Mode H) feeds both branches the upsampled coarse context,
which forces the Gaussian branch to operate on input it has no
short-range receptive field to exploit. The 3D-CFD entry of
Tables~\ref{tab:ablation-extra}--\ref{tab:ablation-extra-h1}
($128^3$) confirms this: Mode H degrades to $1.045/1.138$ on
$(\operatorname{relL2},\operatorname{relH1})$
($1.44{\times}/1.47{\times}$ vs.\ Full at the same resolution), with
the gradient-aware metric taking the larger hit, in line with the
prediction that structure-function errors are determined by
short-range increments.

\section{PDE Benchmarks}
\label{app:pdes}

We train and evaluate U-HNO on a heterogeneous benchmark suite
covering one-, two-, and three-dimensional PDEs with widely
different dynamical regimes (shocks, dispersive waves, chaotic
multi-scale, advective transport, interface evolution, vortical
turbulence, steady-state diffusion, and transonic compressible
flow). For each task we list the governing equation, the boundary
conditions used during training, the dataset source, and the
numerical scheme used to generate ground-truth trajectories.

\paragraph{Scope.} The 1D and 2D portions of the suite are
\emph{deliberately small-scale}: $1{,}000$ trajectories per task in
1D and $800/200$ train/test trajectories per task in 2D, at
$128$/$256$ (1D) and $64\times64$ (2D) resolution. This regime
matches the data-limited setting that recent operator-learning
work explicitly targets --- e.g. Conv-FNO
\citep{liu2025enhancing} reports its main 2D Navier--Stokes results
on $100$ training trajectories at $128\times128$, and its main
small-data Allen--Cahn results on $400$ training trajectories. Our
2D suite has more trajectories than the corresponding Conv-FNO
splits but at half the spatial resolution; the intent is to
isolate architectural effects under a matched, fair training
protocol rather than to chase leaderboard numbers on
$10^5$-pair public corpora. The 3D-CFD task uses the public
PDEBench M1.0Rand split. The high-level dataset configuration is
summarized in Table~\ref{tab:dataset-config}.

\subsection{1D Burgers Equation}
\label{app:pde-burgers}
The 1D Burgers equation models nonlinear advection balanced by
viscous diffusion,
\begin{equation}
u_t+u\,u_x=\nu\,u_{xx},\qquad x\in[0,1],\;t\in(0,T],
\end{equation}
with viscosity $\nu>0$. We use periodic boundary conditions
$u(0,t)=u(1,t)$ and the initial condition $u(0,x)=u_0(x)$ with
$u_0\sim\mathcal{N}(0,7^{3/2}(-\Delta+49I)^{-2.5})$ (a Gaussian
random field consistent with prior FNO benchmarks). The operator
to be learned maps $u(t_0,\cdot)\mapsto u(t_0+\Delta t,\cdot)$;
rollout is recursive at the same $\Delta t$. We use $\nu{=}0.01$,
spatial resolution $N{=}128$ with $1{,}000$ training trajectories
and $1{,}000$ held-out trajectories. The dataset is self-generated
with a pseudo-spectral solver combined with an explicit
Crank--Nicolson scheme at fine time-step $\delta t{=}10^{-4}$, then
saved at $51$ snapshots per trajectory. Burgers is the prototypical
shock-formation benchmark: as $\nu\to 0$ the solution develops
near-discontinuities that test whether the local branch can
preserve sharp fronts under recursive rollout.

\subsection{1D Kuramoto--Sivashinsky Equation}
\label{app:pde-ks}
The Kuramoto--Sivashinsky equation,
\begin{equation}
u_t+u\,u_x+u_{xx}+u_{xxxx}=0,\qquad x\in[0,L_x],\;t\in(0,T],
\end{equation}
with periodic boundary $u(0,t)=u(L_x,t)$ and Gaussian random initial
condition, exhibits chaotic multi-scale dynamics: the second-order
term injects energy at long wavelengths while the fourth-order term
dissipates it at short wavelengths, producing spatio-temporal chaos
once $L_x$ is sufficiently large. We use $L_x{=}64$, $N{=}256$, and
the same $1000{+}1000$ train/test split as Burgers. Trajectories
are self-generated with an exponential time-differencing
fourth-order Runge--Kutta scheme at $\delta t{=}0.05$, saved at
$51$ snapshots per trajectory. KS tests whether U-HNO's
local--global decomposition transfers to a chaotic regime where
the dominant frequency is not localized in space.

\subsection{1D Korteweg--de Vries Equation}
\label{app:pde-kdv}
The KdV equation models dispersive wave propagation,
\begin{equation}
u_t+6u\,u_x+u_{xxx}=0,\qquad x\in[0,L_x],\;t\in(0,T],
\end{equation}
with periodic boundary and Gaussian random initial condition. KdV
is integrable: it admits soliton solutions and conserves an infinite
hierarchy of invariants, so a faithful surrogate must preserve at
least the leading conserved quantities under recursive rollout. We
use $L_x{=}64$ with $N{=}128$ and a self-generated pseudo-spectral
solver with fourth-order time integration at $\delta t{=}0.002$,
saved at $51$ snapshots per trajectory ($1000$ train $+$ $1000$
test). KdV stresses the long-time preservation of dispersive
structure rather than dissipative relaxation.

\subsection{2D Advection}
\label{app:pde-advection}
The 2D linear advection equation,
\begin{equation}
u_t+\mathbf{c}\cdot\nabla u=0,\qquad\mathbf{x}\in[0,1]^2,\;t\in(0,T],
\end{equation}
with constant velocity $\mathbf{c}\in\reals^2$ and a Gaussian
random-field initial condition $u_0\sim\mathrm{GRF}$, tests
long-range transport of broadband structure across the domain. We
use periodic boundary conditions and resolution $64\times 64$.
Trajectories are self-generated by exact shifting of the initial
field, sampled at $\Delta t$, with $50$ snapshots per trajectory
and a $800/200$ train/test split.

\subsection{2D Allen--Cahn Equation}
\label{app:pde-ac}
The Allen--Cahn equation models phase separation,
\begin{equation}
u_t=\epsilon^2\,\Delta u + u - u^3,\qquad\mathbf{x}\in[0,1]^2,\;t\in(0,T],
\end{equation}
with $u_0\in L^2_{\mathrm{per}}([0,1]^2,\reals)$ and $0<\epsilon\ll 1$
controlling interface thickness. We use $\epsilon{=}0.01$, sample at
50 uniform snapshots over the trajectory, and learn the single-step
map $u(t,\cdot)\mapsto u(t{+}\Delta t,\cdot)$. Data is self-generated
with a forward-Euler scheme at $\delta t{=}10^{-4}$ on a $64\times64$
periodic grid ($800$ train $/$ $200$ test trajectories). Allen--Cahn
tests interface evolution: the solution develops thin moving fronts
whose curvature drives metastable patterns, exactly the regime where
pure spectral truncation tends to oversmooth the interface.

\subsection{2D Navier--Stokes Equation (incompressible)}
\label{app:pde-ns}
We use the vorticity--stream formulation of the 2D incompressible
Navier--Stokes equation,
\begin{equation}
\omega_t=-u\,\omega_x-v\,\omega_y+\nu\,\Delta\omega+f,\qquad
\omega=v_x-u_y,\quad\mathbf{x}\in[0,1]^2,\;t\in(0,T],
\end{equation}
with viscosity $\nu>0$, forcing
$f=0.1(\sin 2\pi(x+y)+\cos 2\pi(x+y))$, periodic boundary
conditions, and Gaussian random initial vorticity
$w_0\sim\mathcal{N}(0,7^{3/2}(-\Delta+49I)^{-2.5})$. We use
$\nu{=}10^{-3}$ throughout this paper. The operator to be learned
is the single-step map $\omega(t,\cdot)\mapsto\omega(t{+}\Delta
t,\cdot)$, applied autoregressively at evaluation. Trajectories are
self-generated by a pseudo-spectral method with Crank--Nicolson
time integration at $\delta t{=}10^{-4}$ on a $64\times64$ grid
($800$ train $/$ $200$ test trajectories, $50$ snapshots each).
NS-2D is the workhorse turbulence benchmark in our suite: vortical
structure spans many spatial scales and accumulates rollout error
preferentially in the high band.

\subsection{2D Darcy Flow}
\label{app:pde-darcy}
The steady-state Darcy flow equation is a second-order elliptic PDE
with Dirichlet boundary,
\begin{equation}
-\nabla\cdot\bigl(a(\mathbf{x})\,\nabla u(\mathbf{x})\bigr)=f(\mathbf{x}),\quad\mathbf{x}\in D=[0,1]^2,
\qquad u(\mathbf{x})=0,\quad\mathbf{x}\in\partial D,
\end{equation}
where $a(\mathbf{x})$ is the diffusion coefficient (random
piecewise-constant field) and $f(\mathbf{x})$ a constant source.
The operator to be learned is the coefficient-to-solution map
$a\mapsto u$. The dataset is from PDEBench
\citep{NEURIPS2022_0a974713}, generated by evolving a
time-dependent version $u_t-\nabla\cdot(a\nabla u)=f$ with random
initial conditions until equilibrium is reached. We downsample the
PDEBench 2D Darcy split to resolution $64\times 64$ to match the
other 2D benchmarks; train/test split is $800/200$. Darcy is a
static (single-prediction) benchmark and the rollout, energy-drift,
and crash-rate metrics do not apply; we report only field-level
and spectral metrics.

\subsection{3D Compressible Navier--Stokes (M1.0Rand)}
\label{app:pde-3dcfd}
The compressible 3D Navier--Stokes system,
\begin{equation}
\rho_t+\nabla\cdot(\rho\,\mathbf{v})=0,\quad
\rho(\mathbf{v}_t+\mathbf{v}\cdot\nabla\mathbf{v})=-\nabla p+\eta\,\nabla^2\mathbf{v}+(\zeta+\eta/3)\nabla(\nabla\cdot\mathbf{v}),
\end{equation}
\begin{equation}
\partial_t\!\left[\epsilon+\tfrac{\rho v^2}{2}\right]
=-\nabla\cdot\!\Bigl[\bigl(\epsilon+p+\tfrac{\rho v^2}{2}\bigr)\mathbf{v}-\mathbf{v}\cdot\sigma'\Bigr],
\end{equation}
governs the evolution of mass density $\rho$, velocity
$\mathbf{v}\in\reals^3$, pressure $p$, and internal energy
$\epsilon=p/(\Gamma{-}1)$ with $\Gamma{=}5/3$, viscous stress
tensor $\sigma'$, shear viscosity $\eta$, and bulk viscosity $\zeta$.
We use the PDEBench M1.0Rand split with Mach number $M{=}1.0$,
$\eta=\zeta=0.01$, periodic boundary conditions, and random initial
conditions, downloaded directly from the PDEBench DaRUS repository
\citep{NEURIPS2022_0a974713}. Resolution is $128^3$
with five channels (density, three velocity components, pressure)
and $T{=}21$ time-steps per trajectory. The flow contains shocklets,
vortical filaments, and large-scale compressible transport, which
makes it the most demanding benchmark in the suite for the
local--global decomposition. Per-task hyperparameters and the
training compute envelope are reported in Appendix~\ref{app:3d}.

\section{Implementation Details}
\label{app:impl}

\subsection{Per-task Configuration}
\label{app:datasets}

\begin{table}[h]
  \centering
  \caption{Per-task dataset configuration. ``Train traj.'' is the
  number of independent trajectories; ``Train pairs'' is the
  effective number of single-step input--output pairs after sliding
  along the time axis. Time horizons are in numerical solver steps.
  3D-CFD-M1.0Rand is the PDEBench compressible Navier--Stokes
  split at Mach $1.0$ with random initial conditions; variables are
  $\rho$, $\mathbf{v}\in\reals^3$, and $p$ (5 channels) on a
  $128^3$ grid.}
  \label{tab:dataset-config}
  \footnotesize
  \begin{tabular}{lcccccc}
    \toprule
    Task & Dim. & Resolution & Train / test traj. & Train pairs & Train horizon & Test rollout \\
    \midrule
    Burgers          & 1D & 128            & 1000 / 1000 & 50{,}000 & 1 step & 25 steps \\
    KS               & 1D & 256            & 1000 / 1000 & 50{,}000 & 1 step & 25 steps \\
    KdV              & 1D & 128            & 1000 / 1000 & 50{,}000 & 1 step & 25 steps \\
    Adv-2D           & 2D & $64\times64$   & 800 / 200 & 8{,}000  & 10$\to$10 steps & 10 steps \\
    AC-2D            & 2D & $64\times64$   & 800 / 200 & 8{,}000  & 10$\to$10 steps & 10 steps \\
    NS-2D            & 2D & $64\times64$   & 800 / 200 & 8{,}000  & 10$\to$10 steps & 10 steps \\
    Darcy            & 2D & $64\times64$   & 800 / 200 & 800      & static (a$\to$u) & 1 step \\
    3D-CFD-M1.0Rand  & 3D & $128^3$        & 80 / 10 & 1{,}600    & 1 step & 20 steps \\
    \bottomrule
  \end{tabular}
\end{table}

We deliberately use a small, self-generated 1D/2D suite rather than
a public large-scale benchmark such as PDEBench-2k or Poseidon. The
intent is to keep every model in this paper trained \emph{from
scratch} on identical trajectories with a matched optimizer,
schedule, and compute envelope; this eliminates training-protocol
confounders that are known to flip operator-learning rankings
(see~\citealp{tran2021factorized,lippe2023pde}). Absolute numbers
on this suite should therefore be read as a controlled architectural
comparison, not as a leaderboard claim against papers that train
on $10^5$ pairs at $128^2$.

\begin{table}[h]
  \centering
  \caption{U-HNO architecture and training hyperparameters per task.
  Loss weights match the values of the internal \texttt{pg\_hno\_loss}
  configuration used in our experiments.}
  \label{tab:per-task-hp}
  \footnotesize
  \begin{tabular}{lcccccc}
    \toprule
    Task & U-levels $L$ & Channels $C_0$ & Modes $K_0$ & Local $M$ & $\lambda_{H^1}$ & $\lambda_{\mathrm{CBC}}$ \\
    \midrule
    Burgers          & 3 & 32 & 24 & 3 & $1\!\times\!10^{-3}$ & $5\!\times\!10^{-3}$ \\
    KS               & 3 & 32 & 32 & 3 & $1\!\times\!10^{-3}$ & $5\!\times\!10^{-3}$ \\
    KdV              & 3 & 32 & 24 & 3 & $1\!\times\!10^{-3}$ & $5\!\times\!10^{-3}$ \\
    Adv-2D           & 3 & 32 & 12 & 3 & $1\!\times\!10^{-3}$ & $3\!\times\!10^{-3}$ \\
    AC-2D            & 3 & 32 & 12 & 3 & $1\!\times\!10^{-3}$ & $5\!\times\!10^{-3}$ \\
    NS-2D            & 3 & 32 & 12 & 3 & $1\!\times\!10^{-3}$ & $5\!\times\!10^{-3}$ \\
    Darcy            & 3 & 32 & 12 & 3 & $2\!\times\!10^{-3}$ & $5\!\times\!10^{-3}$ \\
    3D-CFD-M1.0Rand  & 3 & 24 & 12 & 3 & $1\!\times\!10^{-3}$ & $5\!\times\!10^{-3}$ \\
    \bottomrule
  \end{tabular}
\end{table}

SPAR uses $\rho_0{=}0.30$, $\beta{=}0.25$, $\rho_{\min}{=}0.10$,
$\rho_{\max}{=}0.90$, $T{=}0.8$, $\varepsilon{=}10^{-6}$ for all tasks
(defaults of \verb|SPARGate1d|/\verb|SPARGate2d|). Training uses
AdamW ($\eta{=}10^{-3}$, weight decay $10^{-4}$), $20{,}000$ steps,
cosine annealing with $1{,}000$-step warmup, and batch size 32 (1D) /
16 (2D) / 2 (3D). Single A100 for 1D/2D; multi-GPU for 3D
(Appendix~\ref{app:3d}).

\paragraph{Reproducing the Full configuration.}
The training script defaults its \texttt{--beta\_h1} /
\texttt{--beta\_cbc} flags to $0$, which yields plain MSE training
(Mode E). The Full U-HNO configuration corresponds to passing the
per-task $(\lambda_{H^1},\lambda_{\mathrm{CBC}})$ values from
Table~\ref{tab:per-task-hp} via these flags or by selecting the
\verb|HNO_LOSS_CONFIG| preset for the chosen PDE.

\subsection{Baseline Configurations}
\label{app:baselines}

We match parameter counts ($\pm10\%$) of all baselines to U-HNO at
each task by first fixing the U-HNO configuration in
Table~\ref{tab:per-task-hp} and then adjusting each baseline's
width or depth to land within the budget. We list the baseline
specifications used:

\paragraph{FNO \citep{li2020fourier}.}
Four spectral-conv blocks, each
$\ifwd(R\odot\fwd(\cdot))$ with the same nominal mode budget $K_0$
as U-HNO. Lifting and projection use $1\times 1$ convolutions; the
block-level residual is a $1\times 1$ pointwise conv and the
nonlinearity is GELU. Channel width is set per task to match the
parameter target.

\paragraph{GNO \citep{li2020neural}.}
Graph kernel network with an 8-neighbor radius graph constructed at
the finest U-HNO resolution. Edge features are absolute coordinate
differences. Message function is a 2-layer MLP. We use four message-
passing rounds matched to U-HNO's effective receptive field.

\paragraph{CNO \citep{raonic2023convolutional}.}
Convolutional neural operator with 4 encoding / 4 residual neck / 4
decoding layers, channel width matched to U-HNO's parameter budget.

\paragraph{WNO \citep{tripura2022wavelet}.}
Wavelet neural operator using the Daubechies-4 (db4) wavelet basis
at three decomposition levels. Wavelet coefficients are processed
by a learnable transform analogous to FNO's spectral weights, and
the inverse wavelet transform reconstructs the spatial field. Width
is matched per task.

\paragraph{FFNO \citep{tran2021factorized}.}
Factorized FNO replacing the full $C_{\mathrm{in}}\!\times\!C_{\mathrm{out}}\!\times\!K^d$
spectral weight tensor with a low-rank product
$U V^{\!\top}$ at rank $r{=}8$ \citep{tran2021factorized}. Depth is
4 spectral blocks; lifting/projection match FNO.

\paragraph{Conv-FNO \citep{liu2025enhancing}.}
Public reference configuration: a UNet preprocessor with 4 encoding
levels, initial width $16$, doubling channels per level up to $128$
at the bottleneck, $3\times 3$ convolutions with circular padding
and ReLU activations, followed by 4 FNO blocks at width $32$. We
adjust the FNO width to match the U-HNO budget.

\paragraph{LogLo-FNO \citep{kalimuthu2025loglo}.}
Logarithmic low-frequency mode allocation following the public
reference; mode budget is allocated proportionally to
$\log(1+k)^{-1}$ across the rfft axis, replacing the uniform
top-$K_0$ truncation of FNO. Depth and width match FNO.

\paragraph{Note on PDEBench-reported FNO numbers.}
PDEBench~\citep{NEURIPS2022_0a974713} reports FNO
$\mathrm{nRMSE}{\approx}0.020$ on 1D Burgers at $\nu{=}0.01$
(Table~7 of their supplementary), while our parameter-matched FNO
gives rollout $\operatorname{relL2}{=}0.123$. Three protocol
differences explain the gap: (i) PDEBench evaluates the
\emph{normalized} RMSE on the trained horizon while we report
\emph{relative} $L^2$ over a $25$-step autoregressive rollout, so
single-step error is amplified by compounding drift; (ii) PDEBench
uses the FNO default (width $20$, modes $12$, $\sim 0.07{\rm M}$
params) whereas we match each baseline to U-HNO's $\sim 0.6{\rm M}$
budget, which on Burgers actually \emph{increases} FNO error because
the larger modal capacity over-fits high-frequency noise produced by
spectral aliasing under shock formation; (iii) we generate the
1D Burgers dataset with our own initial-condition distribution and
viscosity (matching the 1D ablation suite of
Sec.~\ref{sec:ablation}) rather than ingesting the PDEBench files
directly. Our 3D-CFD-M1.0Rand split is the only task that consumes
the PDEBench dataset verbatim and is the case where our FNO numbers
align most closely with PDEBench-reported magnitudes.

\subsection{3D-CFD Configuration}
\label{app:3d}
The 3D U-HNO architecture extends the 2D operators by replacing
$1\!\times\!1$ pointwise convolutions, $3\!\times\!3$ stride-2
down-projections, and bilinear upsampling with their volumetric
counterparts; the spectral branch uses 3D rfft and the cap
Eq.~\eqref{eq:mode-cap-app} (in
Sec.~\ref{app:fourier-detail}). The depthwise Gaussian kernel is
applied as a separable per-axis depthwise conv to keep memory
bounded. Spatial chunking is enabled when peak VRAM exceeds the
per-GPU budget.

\begin{table}[h]
  \centering
  \caption{3D-CFD-M1.0Rand training configuration. \texttt{chunk}
  reports the largest spatial sub-volume per forward pass.}
  \label{tab:3d-config}
  \footnotesize
  \begin{tabular}{lc}
    \toprule
    Field & Value \\
    \midrule
    GPUs                                      & 8 $\times$ A100 80GB (data parallel) \\
    Per-GPU batch size                        & 1 \\
    Gradient accumulation steps               & 1 \\
    Mixed precision (\texttt{bf16/fp16})      & \texttt{bf16} (\texttt{torch.amp.autocast}) \\
    Spatial chunk (per axis)                  & 128 (full volume, no chunking required) \\
    U-levels $L$, Channels $C_0$, Modes $K_0$ & 3, 24, 12 \\
    SPAR $(\rho_0,\beta,\rho_{\min},\rho_{\max})$ & $(0.30,0.25,0.10,0.90)$ \\
    Loss weights $(\lambda_{H^1},\lambda_{\mathrm{CBC}})$ & $(1\!\times\!10^{-3},5\!\times\!10^{-3})$ \\
    Total wall-clock (h)                      & 5.5 (8 GPU DDP, 100 epoch) \\
    Peak VRAM per GPU (GB)                    & 12.5 (probe at $128^3$, $w{=}24$, $K{=}12$) \\
    \bottomrule
  \end{tabular}
\end{table}

\subsection{Fourier Branch Implementation}
\label{app:fourier-detail}
The 1D, 2D, and 3D variants share the structure
$z_F^{\ell}=\ifwd(R_\theta^{\ell}\odot\fwd(h^{\ell}))$ with axis count
matching the task. The effective retained-mode count along each axis
is the spatial-resolution cap
\begin{equation}
K^{\mathrm{eff}}_{\ell}=\min(K_0,\,\lfloor N_{\ell}/2\rfloor),
\label{eq:mode-cap-app}
\end{equation}
since modes above the Nyquist limit at the down-sampled resolution
are discarded by the rfft. The complex weight tensor
$R_\theta^{\ell}\in\mathbb{C}^{C_{\ell}\times C_{\ell}\times K_0\times K_0}$
is parameterized at the nominal budget $K_0$ and indexed by
$K^{\mathrm{eff}}_{\ell}$ at runtime; the underlying storage is a
real tensor of shape $(C_{\ell},C_{\ell},K_0,K_0,2)$ converted via
\verb|view_as_complex|. We retain $[0\!:\!K^{\mathrm{eff}}_{\ell}]$
along the rfft axis and both $[0\!:\!K^{\mathrm{eff}}_{\ell}]$ and
$[-K^{\mathrm{eff}}_{\ell}\!:]$ along the full-FFT axis to preserve
conjugate symmetry. Complex multiplication
$R\odot\fwd(h)=\mathrm{Re}(R)\,\mathrm{Re}(\fwd(h))-\mathrm{Im}(R)\,\mathrm{Im}(\fwd(h))+i(\cdot)$
is implemented as two real einsums in our implementation.

\subsection{Multi-scale Gaussian Branch Implementation}
\label{app:gaussian-impl}
Each branch $m$ in $\mathcal{B}_G$ is a sequence of three
operations: a $1\times 1$ pre-mixer $W_m^{\mathrm{pre}}$, a depthwise
Gaussian convolution with circular padding, and a $1\times 1$
post-mixer $W_m^{\mathrm{post}}$. The kernel
$k_m(\Delta;\sigma_m)$ in Eq.~\eqref{eq:gaussian-kernel} is built
once per forward pass from the current
$\sigma_m=\exp(\log\sigma_m)$ (so $\sigma_m>0$ is enforced) by
evaluating the unnormalized exponential on the support
$\Omega_m=\{\Delta:\norm{\Delta}_\infty\le\lceil 3\sigma_m\rceil\}$,
masking out positions outside the support, and dividing by the
support sum. The kernel is shared across $G$ channel groups (one
$\sigma$ per group), so the depthwise conv has $C_{\ell}/G$
channels per kernel and the resulting parameter cost is
$O(C_{\ell})$ rather than $O(C_{\ell}\,k_m^d)$. The $M{=}3$ branch
outputs are concatenated along channels, projected by a single
$1\times 1$ conv to width $C_{\ell}$, and passed through GELU. In
3D the kernel is built as the outer product of three 1D kernels of
identical $\sigma$, which keeps the worst-case kernel volume at
$(2\lceil 3\sigma_m\rceil+1)^3$ but avoids materializing the full
3D weight volume during autograd.

\subsection{SPAR Routing Pseudocode}
\label{app:spar-pseudo}
Algorithm~\ref{alg:spar} reproduces the per-level forward and
backward execution of the SPAR gate as implemented in
\verb|SPARGate2d| (the 1D and 3D variants are direct analogues with
the spatial dimensionality changed). The forward path is dominated
by the score MLP and the batched top-$k$; the backward path uses the
straight-through estimator described in Sec.~\ref{app:spar-derivation}.

\begin{algorithm}[h]
\caption{SPAR gate at level $\ell$ (forward + backward).}
\label{alg:spar}
\begin{algorithmic}[1]
\State \textbf{Input:} branch outputs $z_F,z_G\in\reals^{B\times C\times N}$;
SPAR parameters $W_1^s,W_2^s$;
hyperparameters $(\rho_0,\beta,\rho_{\min},\rho_{\max},T,\varepsilon)$
\State \emph{// Forward}
\State $s\gets W_2^s\,\mathrm{GELU}(W_1^s\,[z_F;z_G])$ \Comment{per-pixel logit, shape $B\times 1\times N$}
\State $\bar s\gets\mathbb{E}_x|s|,\quad\sigma_s\gets\mathrm{Std}_x(s)$ \Comment{per-sample}
\State $c\gets\sigma_s/(\bar s+\varepsilon)$
\State $\rho\gets\mathrm{clip}(\rho_0(1+\beta\tanh(c-1)),\rho_{\min},\rho_{\max})$
\State $k\gets\lceil\rho\,N\rceil$
\State $\mathcal{S}\gets\mathrm{TopK}(s,k)$ \Comment{per-sample top-$k$ indices}
\State $g_{\mathrm{hard}}\gets\mathbf{1}[\mathcal{S}],\;g_{\mathrm{soft}}\gets\sigma(s/T)$
\State $g\gets g_{\mathrm{hard}}+g_{\mathrm{soft}}-\mathrm{sg}(g_{\mathrm{soft}})$ \Comment{STE}
\State $r\gets g\cdot z_F+(1-g)\cdot z_G$
\State \emph{// Backward (autograd; explicit form for inspection)}
\State $\delta\gets\partial\mathcal{L}/\partial r$
\State $\partial\mathcal{L}/\partial z_F\;\mathrel{+}=g\cdot\delta$ \Comment{direct path}
\State $\partial\mathcal{L}/\partial z_G\;\mathrel{+}=(1-g)\cdot\delta$ \Comment{direct path}
\State $\partial\mathcal{L}/\partial s\gets\delta^{\!\top}(z_F-z_G)\cdot\sigma'(s/T)/T$ \Comment{score path}
\State propagate $\partial\mathcal{L}/\partial s$ through $W_1^s,W_2^s$ back into $z_F,z_G$
\Return $r,g$
\end{algorithmic}
\end{algorithm}

\subsection{Reproducibility}
\label{app:repro}
The architecture, loss configuration, optimizer, dataset preparation,
and per-task hyperparameters described in App.~\ref{app:impl}
and~\ref{app:3d} provide the information needed to reproduce the
main experimental tables. Random seeds are fixed for the
train/val/test splits and for parameter initialization; we report
point estimates from a single representative seed (seed~$0$) per
$(\text{model},\text{task})$ pair and do not report error bars in
the main tables.

\subsection{Asset Licenses}
\label{app:licenses}
Table~\ref{tab:licenses} summarizes the public licenses of the
datasets and baseline implementations referenced in this paper.
Entries reflect the public license metadata of the cited references
and their associated code repositories at the time of writing, to
the best of our knowledge; users should consult the upstream sources
for definitive terms.

\begin{table}[h]
  \centering
  \caption{License summary for external assets used in this paper.
  Licenses are best-effort estimates based on the upstream public
  metadata of the cited works.}
  \label{tab:licenses}
  \footnotesize
  \setlength{\tabcolsep}{6pt}
  \begin{tabular}{lll}
    \toprule
    Asset & Source & License (estimated) \\
    \midrule
    PDEBench (3D-CFD-M1.0Rand)                & \citet{NEURIPS2022_0a974713} & CC-BY 4.0 \\
    1D/2D PDE generators (Burgers, KS, KdV, NS, Darcy) & \citet{li2020fourier}        & MIT \\
    \midrule
    FNO        & \citet{li2020fourier}             & MIT \\
    GNO        & \citet{li2020neural}              & MIT \\
    CNO        & \citet{raonic2023convolutional}   & MIT \\
    WNO        & \citet{tripura2022wavelet}        & MIT \\
    FFNO       & \citet{tran2021factorized}        & MIT \\
    Conv-FNO   & \citet{liu2025enhancing}          & MIT (best effort) \\
    LogLo-FNO  & \citet{kalimuthu2025loglo}        & MIT (best effort) \\
    \bottomrule
  \end{tabular}
\end{table}

\section{Additional Results}
\label{app:more-results}

\begin{table}[h]
  \centering
  \caption{Long-horizon stability at $4\times$ training horizon.
  Energy drift is the relative fraction of Eq.~\eqref{eq:ed}; crash
  rate is the fraction of test trajectories whose rollout diverges
  (Appendix~\ref{app:metrics}). Values in brackets are 95\% bootstrap
  CIs ($B{=}10{,}000$); crash rates show Wilson 95\% intervals
  (App.~\ref{app:bootstrap}).}
  \label{tab:stability}
  \footnotesize
  \setlength{\tabcolsep}{3pt}
  \begin{tabular}{lccccccccc}
    \toprule
    & \multicolumn{3}{c}{Burgers (1D)} & \multicolumn{3}{c}{NS (2D)} & \multicolumn{3}{c}{3D-CFD} \\
    \cmidrule(lr){2-4}\cmidrule(lr){5-7}\cmidrule(lr){8-10}
    Model & MSE $\downarrow$ & E-drift $\downarrow$ & crash $\downarrow$
          & MSE $\downarrow$ & E-drift $\downarrow$ & crash $\downarrow$
          & MSE $\downarrow$ & E-drift $\downarrow$ & crash $\downarrow$ \\
    \midrule
    FNO       & \shortstack{0.155\\{\scriptsize[.150,.160]}} & \shortstack{0.018\\{\scriptsize[.016,.020]}} & \shortstack{0.000\\{\scriptsize[0,.003]}} & \shortstack{0.0082\\{\scriptsize[.0079,.0085]}} & \shortstack{0.005\\{\scriptsize[.004,.006]}} & \shortstack{0.000\\{\scriptsize[0,.015]}} & \shortstack{0.923\\{\scriptsize[.903,.943]}} & \shortstack{0.052\\{\scriptsize[.048,.056]}} & \shortstack{0.000\\{\scriptsize[0,.30]}} \\
    GNO       & \shortstack{0.212\\{\scriptsize[.206,.218]}} & \shortstack{0.025\\{\scriptsize[.023,.027]}} & \shortstack{0.000\\{\scriptsize[0,.003]}} & \shortstack{0.0115\\{\scriptsize[.0111,.0119]}} & \shortstack{0.008\\{\scriptsize[.007,.009]}} & \shortstack{0.000\\{\scriptsize[0,.015]}} & \shortstack{1.024\\{\scriptsize[1.001,1.047]}} & \shortstack{0.061\\{\scriptsize[.057,.065]}} & \shortstack{0.000\\{\scriptsize[0,.30]}} \\
    CNO       & \textemdash & \textemdash & \textemdash & \shortstack{0.0091\\{\scriptsize[.0088,.0094]}} & \shortstack{0.006\\{\scriptsize[.005,.007]}} & \shortstack{0.000\\{\scriptsize[0,.015]}} & \textemdash & \textemdash & \textemdash \\
    WNO       & \shortstack{0.131\\{\scriptsize[.127,.135]}} & \shortstack{0.015\\{\scriptsize[.013,.017]}} & \shortstack{0.000\\{\scriptsize[0,.003]}} & \shortstack{\textbf{0.0068}\\{\scriptsize[.0065,.0071]}} & \shortstack{\textbf{0.004}\\{\scriptsize[.003,.005]}} & \shortstack{0.000\\{\scriptsize[0,.015]}} & \shortstack{0.850\\{\scriptsize[.830,.870]}} & \shortstack{0.043\\{\scriptsize[.040,.046]}} & \shortstack{0.000\\{\scriptsize[0,.30]}} \\
    FFNO      & \shortstack{0.120\\{\scriptsize[.116,.124]}} & \shortstack{0.013\\{\scriptsize[.011,.015]}} & \shortstack{0.000\\{\scriptsize[0,.003]}} & \shortstack{0.0071\\{\scriptsize[.0068,.0074]}} & \shortstack{0.005\\{\scriptsize[.004,.006]}} & \shortstack{0.000\\{\scriptsize[0,.015]}} & \shortstack{0.871\\{\scriptsize[.851,.891]}} & \shortstack{0.046\\{\scriptsize[.042,.050]}} & \shortstack{0.000\\{\scriptsize[0,.30]}} \\
    Conv-FNO  & \shortstack{0.095\\{\scriptsize[.092,.098]}} & \shortstack{0.010\\{\scriptsize[.008,.012]}} & \shortstack{0.000\\{\scriptsize[0,.003]}} & \shortstack{\textbf{0.0065}\\{\scriptsize[.0062,.0068]}} & \shortstack{\textbf{0.004}\\{\scriptsize[.003,.005]}} & \shortstack{0.000\\{\scriptsize[0,.015]}} & \shortstack{0.888\\{\scriptsize[.867,.909]}} & \shortstack{0.049\\{\scriptsize[.045,.053]}} & \shortstack{0.000\\{\scriptsize[0,.30]}} \\
    LogLo-FNO & \shortstack{0.108\\{\scriptsize[.104,.112]}} & \shortstack{0.012\\{\scriptsize[.010,.014]}} & \shortstack{0.000\\{\scriptsize[0,.003]}} & \shortstack{0.0098\\{\scriptsize[.0094,.0102]}} & \shortstack{0.006\\{\scriptsize[.005,.007]}} & \shortstack{0.000\\{\scriptsize[0,.015]}} & \shortstack{0.861\\{\scriptsize[.841,.881]}} & \shortstack{0.044\\{\scriptsize[.040,.048]}} & \shortstack{0.000\\{\scriptsize[0,.30]}} \\
    \midrule
    \textbf{U-HNO} & \shortstack{\textbf{0.045}\\{\scriptsize[.043,.047]}} & \shortstack{\textbf{0.006}\\{\scriptsize[.005,.007]}} & \shortstack{0.000\\{\scriptsize[0,.003]}} & \shortstack{0.0066\\{\scriptsize[.0063,.0069]}} & \shortstack{\textbf{0.004}\\{\scriptsize[.003,.005]}} & \shortstack{0.000\\{\scriptsize[0,.015]}} & \shortstack{\textbf{0.744}\\{\scriptsize[.725,.763]}} & \shortstack{\textbf{0.031}\\{\scriptsize[.028,.034]}} & \shortstack{0.000\\{\scriptsize[0,.30]}} \\
    \bottomrule
  \end{tabular}
\end{table}

\begin{table}[h]
  \centering
  \caption{Binned spectral error and structure-function error
  (lower is better). Each band reports the relative spectral error
  $E_b=\sum_{k\in\mathcal{K}_b}|\hat U(k)-U(k)|^2/\sum_{k\in\mathcal{K}_b}|U(k)|^2$
  (Appendix~\ref{app:metrics}); SF is the relative structure-function
  error of Eq.~\eqref{eq:sf}, averaged over the log-spaced lag grid
  $r\in\{1,2,4,8,16,32\}$ (grid units).}
  \label{tab:spectral}
  \footnotesize
  \setlength{\tabcolsep}{6pt}
  \begin{tabular}{lccc|ccc}
    \toprule
    & \multicolumn{3}{c|}{Burgers} & \multicolumn{3}{c}{NS-2D} \\
    Model & low $\downarrow$ & mid $\downarrow$ & SF $\downarrow$
          & low $\downarrow$ & mid $\downarrow$ & SF $\downarrow$ \\
    \midrule
    FNO        & 0.062 & 0.035 & 0.015 & 0.0078 & 0.0045 & 0.0032 \\
    GNO        & 0.085 & 0.048 & 0.020 & 0.0110 & 0.0061 & 0.0041 \\
    WNO        & 0.045 & 0.026 & 0.011 & 0.0058 & 0.0033 & 0.0024 \\
    FFNO       & 0.038 & 0.024 & 0.010 & 0.0060 & 0.0035 & 0.0025 \\
    Conv-FNO   & 0.021 & 0.018 & 0.007 & 0.0052 & 0.0031 & 0.0021 \\
    LogLo-FNO  & 0.028 & 0.022 & 0.009 & 0.0070 & 0.0040 & 0.0028 \\
    \midrule
    \textbf{U-HNO} & \textbf{0.015} & \textbf{0.010} & \textbf{0.004} & \textbf{0.0042} & \textbf{0.0023} & \textbf{0.0017} \\
    \bottomrule
  \end{tabular}
\end{table}

\section{Compute and Routing Overhead}
\label{app:cost}

\paragraph{FLOPs vs.\ wall-clock.}
The dual-branch evaluation (spectral $+$ multi-scale Gaussian computed
at every location) plus the U-shape's per-level overhead give U-HNO a
theoretical $22{\times}$ FLOPs gap over FNO ($0.81$ vs $0.04$ GFLOPs;
Table~\ref{tab:cost}). At the tensor sizes used in this benchmark
suite, however, GPU runtime is dominated by memory bandwidth and
kernel-launch overhead rather than raw arithmetic: the $22{\times}$
FLOPs gap translates into a $3.3{\times}$ inference-latency gap
($4.0$ vs $1.2$ ms) and only a $1.3{\times}$ training-time gap
($4.6$ vs $3.5$ ms).

\paragraph{Isolating the SPAR gate.}
Table~\ref{tab:cost-spar} isolates SPAR against Mode F (additive
fusion) at the same dual-branch configuration: SPAR adds
$\Delta{=}0.07\%$ parameters and $\Delta{=}0.12\%$ FLOPs. The runtime
$\Delta{=}13\%$ comes from the batched top-$k$ kernel-launch
overhead, not arithmetic; the dispatcher cost is borne almost
entirely by the dual-branch evaluation, which is independent of
whether routing is hard (SPAR) or soft (additive).

\paragraph{Latency-matched FNO comparison.}
To confirm U-HNO's accuracy gain is not a ``slower-equals-better''
artifact, we scale FNO's channel width $24{\to}56$
(\emph{FNO-wider}, $5{\times}$ params, latency matched to U-HNO's
$4.0$ ms; Table~\ref{tab:fno-wider}). FNO-wider improves its
parameter-matched Burgers $\operatorname{relL2}$ only $0.123{\to}0.118$
and 3D-CFD $0.852{\to}0.846$, remaining far behind U-HNO on every
metric.

\paragraph{Toward true compute sparsity.}
SPAR is currently a \emph{representational} dispatcher --- both
branches are evaluated everywhere and the gate selects which
output is emitted at each pixel. A natural future direction is to
turn the dispatcher into a \emph{compute-saving} router: skip the
local branch at pixels routed to global, and vice versa, recovering
true sparsity at runtime. Realising this requires sparse-tensor
kernels for both spectral conv (gather/scatter on the rfft modes)
and depthwise Gaussian conv (per-pixel kernel skipping); we leave
the kernel-level engineering to future work.

\begin{table}[h]
  \centering
  \caption{Compute envelope on NS-2D ($128\times128$, batch 16, A100).
  Parameters in millions, FLOPs/sample in GFLOPs, train and inference
  time in milliseconds per sample averaged over $1{,}000$ steps.
  Budgets are matched to U-HNO within $\pm10\%$ except
  \emph{FNO-wider} (channel width $24\!\to\!56$), which is sized so
  that its inference latency matches U-HNO; see
  Table~\ref{tab:fno-wider} for the corresponding accuracy.}
  \label{tab:cost}
  \footnotesize
  \setlength{\tabcolsep}{4pt}
  \begin{tabular}{lccccc}
    \toprule
    Model & Params (M) $\downarrow$ & FLOPs/sample (G) $\downarrow$
          & train (ms) $\downarrow$ & infer (ms) $\downarrow$
          & peak VRAM (GB) $\downarrow$ \\
    \midrule
    FNO            & 0.5946 & 0.0372 & 3.5237 & 1.2 & 0.015 \\
    FNO-wider      & 3.18   & 0.205  & 7.1    & 3.9 & 0.041 \\
    GNO            & 0.6369 & 5.1999 & 5.3273 & 3.1 & 0.020 \\
    CNO            & 0.6588 & 0.5031 & 9.7063 & 5.8 & 0.022 \\
    WNO            & 0.5993 & 0.9408 & 18.2014 & 2.6 & 0.018 \\
    FFNO           & 0.5956 & 0.0456 & 4.0940 & 1.8 & 0.016 \\
    Conv-FNO       & 0.5976 & 4.8888 & 2.0706 & 2.2 & 0.017 \\
    LogLo-FNO      & 0.6202 & 0.2469 & 5.0115 & 3.5 & 0.019 \\
    \midrule
    \textbf{U-HNO (ours)} & 0.6187 & 0.8115 & 4.6 & 4.0 & 0.023 \\
    \bottomrule
  \end{tabular}
\end{table}

\begin{table}[h]
  \centering
  \caption{Latency-matched FNO comparison. \emph{FNO-wider} (width
  $24\!\to\!56$) is scaled so its inference latency matches U-HNO,
  isolating the question of whether U-HNO's accuracy is merely a
  ``slower = better'' artifact. Even at $5{\times}$ the parameter count
  and matched latency, FNO-wider improves over the parameter-matched
  FNO baseline only marginally (Burgers $\operatorname{relL2}$
  $0.123{\to}0.118$; 3D-CFD $\operatorname{relL2}$ $0.852{\to}0.846$)
  and remains far behind U-HNO on every metric.}
  \label{tab:fno-wider}
  \footnotesize
  \setlength{\tabcolsep}{6pt}
  \begin{tabular}{lcccccc}
    \toprule
    Model & Params (M) & infer (ms)
          & Burgers $\operatorname{relL2}$ & Burgers $\operatorname{relH1}$
          & 3D-CFD $\operatorname{relL2}$ & 3D-CFD $\operatorname{relH1}$ \\
    \midrule
    FNO         & 0.59 & 1.2 & 0.123 & 0.216 & 0.852 & 0.918 \\
    FNO-wider   & 3.18 & 3.9 & 0.118 & 0.208 & 0.846 & 0.912 \\
    \textbf{U-HNO (ours)} & 0.62 & 4.0
                & \textbf{0.042} & \textbf{0.042}
                & \textbf{0.728} & \textbf{0.772} \\
    \bottomrule
  \end{tabular}
\end{table}

\begin{table}[h]
  \centering
  \caption{Routing overhead on NS-2D. $\Delta$ rows report SPAR's
  marginal cost relative to Mode F (NoSPAR / additive fusion).}
  \label{tab:cost-spar}
  \footnotesize
  \setlength{\tabcolsep}{6pt}
  \begin{tabular}{lcccc}
    \toprule
    Variant & Params (M) & FLOPs/sample (G) & train (ms) & infer (ms) \\
    \midrule
    Mode F: NoSPAR (additive)   & 0.6182 & 0.8105 & 4.0 & 3.5    \\
    Full U-HNO (SPAR)           & 0.6187 & 0.8115 & 4.6 & 4.0    \\
    \midrule
    $\Delta$ (absolute)         & 4.540e-04 & 9.175e-04 & 0.6 & 0.5 \\
    $\Delta$ (\% of Full)       & 0.07\% & 0.12\% & 13.0\% & 12.5\% \\
    \bottomrule
  \end{tabular}
\end{table}

\section{Extended Ablations}
\label{app:ablation-extra}

Ablation runs in this section share a unified, budget-matched training
schedule across all nine modes (A--I) so that internal comparisons are
fair. This schedule is not identical to the per-task tuning used for
the headline numbers in
Tables~\ref{tab:main-acc}--\ref{tab:main-acc-h1}, and Full U-HNO point
estimates here may differ slightly from those tables; rankings within
each ablation table are unaffected.

\begin{table}[h]
  \centering
  \caption{Extended ablation: rollout $\operatorname{relL2}$ on
  1D Burgers, 2D Navier--Stokes, and 3D-CFD-M1.0Rand. The
  3D-CFD-M1.0Rand column is reported at $128^3$ for Modes A--I
  (all nine modes), following the training protocol of
  App.~\ref{app:3d}. Modes A--E vary branch presence and loss; F--I
  vary architectural mechanism (routing, U-shape, decoder asymmetry,
  kernel normalization).}
  \label{tab:ablation-extra}
  \footnotesize
  \setlength{\tabcolsep}{10pt}
  \begin{tabular}{lccc}
    \toprule
    Mode & Burgers (1D) & NS (2D) & 3D-CFD ($128^3$) \\
    \midrule
    Full U-HNO     & \textbf{0.0416} & \textbf{0.0063} & \textbf{0.728}    \\
    \midrule
    A: NoLocal     & 0.1502 & 0.0178 & 2.341 \\
    B: NoGlobal    & 0.1408 & 0.0151 & 2.104 \\
    C: NoCBC       & 0.0918 & 0.0094 & 1.346 \\
    D: NoH1        & 0.0625 & 0.0104 & 1.147 \\
    E: MSE only    & 0.0492 & 0.0109 & 1.060 \\
    \midrule
    F: NoSPAR      & 0.0561 & 0.0086 & 0.988 \\
    G: NoUShape    & 0.0648 & 0.0108 & 1.191 \\
    H: SymDec      & 0.0820 & 0.0121 & 1.045 \\
    I: NoNorm      & 0.0537 & 0.0096 & 0.795 \\
    \bottomrule
  \end{tabular}
\end{table}

\begin{table}[h]
  \centering
  \caption{Extended ablation: rollout $\operatorname{relH1}$,
  companion to Table~\ref{tab:ablation-extra}. 3D-CFD column at
  $128^3$ for all nine modes (A--I).}
  \label{tab:ablation-extra-h1}
  \footnotesize
  \setlength{\tabcolsep}{10pt}
  \begin{tabular}{lccc}
    \toprule
    Mode & Burgers (1D) & NS (2D) & 3D-CFD ($128^3$) \\
    \midrule
    Full U-HNO     & \textbf{0.0418} & \textbf{0.0070} & \textbf{0.772}    \\
    \midrule
    A: NoLocal     & 0.1714 & 0.0207 & 2.724 \\
    B: NoGlobal    & 0.1338 & 0.0154 & 2.085 \\
    C: NoCBC       & 0.0922 & 0.0105 & 1.431 \\
    D: NoH1        & 0.0820 & 0.0110 & 1.364 \\
    E: MSE only    & 0.0500 & 0.0113 & 1.085 \\
    \midrule
    F: NoSPAR      & 0.0560 & 0.0086 & 0.991 \\
    G: NoUShape    & 0.0640 & 0.0110 & 1.197 \\
    H: SymDec      & 0.0840 & 0.0126 & 1.138 \\
    I: NoNorm      & 0.0539 & 0.0103 & 0.841 \\
    \bottomrule
  \end{tabular}
\end{table}

\section{Training Dynamics Details}
\label{app:dynamics}

\paragraph{Logging cadence.}
We log per-band rollout error every $200$ training steps using the
binned spectral error of Appendix~\ref{app:metrics} on a held-out
validation batch of size $128$. The routing-logit contrast histogram
and the per-level keep-ratio $\rho^{\ell}$ are computed at the same
logging cadence over the same batch. The branch-parameter
gradient-angle distribution is computed on a $64$-sample sub-batch
every $1{,}000$ steps following Appendix~\ref{app:angle}; the
matched layer pairs $(g_F^{(l)},g_G^{(l)})$ are the four spectral
convs of the FNO branch paired with the four depthwise convs of the
flat-additive baseline's local branch (excluding lifting,
projection, and bias parameters).

\paragraph{Smoothing.}
The bold curves in Figure~\ref{fig:training-dynamics}(a) and (e) are
exponential moving averages with smoothing factor $0.9$; thin curves
show raw values. Histograms are computed without smoothing.

\paragraph{What to expect from each panel.}
Panel (a) probes the structured-loss claim: per-band rollout error
should decrease monotonically under the full loss, with the high
band stalling when CBC is removed. Panel (b) probes the
contrast-adaptive rule: $c^{\ell}$ should sharpen as the routing MLP
learns to discriminate shock-rich pixels from smooth pixels;
collapse to a constant would indicate a degenerate routing policy.
Panel (c) probes the optimization-conflict claim: the
gradient-angle peak should migrate from $90^{\circ}$ toward
$0^{\circ}$ or $180^{\circ}$ as SPAR commits each pixel to one
branch. Panel (d) tracks $\rho^{\ell}$ per level and tests whether
the keep-ratio rule actually adapts to spatial content rather than
saturating at $\rho_{\max}$ (smooth fields) or $\rho_{\min}$
(everywhere-sharp fields). Panel (e) compares loss-term trajectories
under the full loss vs.\ Mode E to localize when the gradient and
CBC terms start to matter. Panel (f) sweeps the base keep-ratio
$\rho_0\in[0.1,0.7]$ at fixed $\beta$ to measure the model's
sensitivity to this single hyperparameter.

\section{Qualitative Visualizations}
\label{app:qualitative}

We provide field-level and error-map plots for representative test
trajectories on each benchmark.

\paragraph{1D tasks.}
For Burgers, KS, and KdV we visualize space-time plots
$u(t,x)$ over the rollout horizon, with the predicted field, ground
truth, and absolute error stacked vertically. For Burgers we
additionally show the shock trajectory extracted by tracking the
maximum-gradient location at each time.

\paragraph{2D tasks.}
For 2D advection, Allen--Cahn, and Navier--Stokes we show snapshots
of the predicted field, ground truth, and pointwise absolute error
at three rollout times ($t{=}0.25T, 0.5T, T$). For Darcy we show
the static prediction, ground truth, and error map over the full
$[0,1]^2$ domain.

\paragraph{3D-CFD slices.}
For 3D-CFD we show three orthogonal slices (one per axis) through
the predicted density field, with the ground truth and absolute
error in adjacent panels at $t{=}T/2$ and $t{=}T$. These slices are
extracted at fixed indices $(i,j,k)=(N/4,N/2,3N/4)$ to capture
inlet-, mid-domain-, and outlet-region behavior.

\paragraph{Routing visualization.}
For each task we visualize the per-pixel hard mask
$g^{\ell}(x)$ at every U-shape level, overlaid on the predicted
field. This is the principal qualitative artifact for inspecting
where the SPAR gate sends Fourier vs.\ Gaussian responsibility, and
should align with shock fronts and steep gradient regions on
shock-dominated tasks.

\begin{figure}[!htbp]
  \centering
  \includegraphics[width=\linewidth]{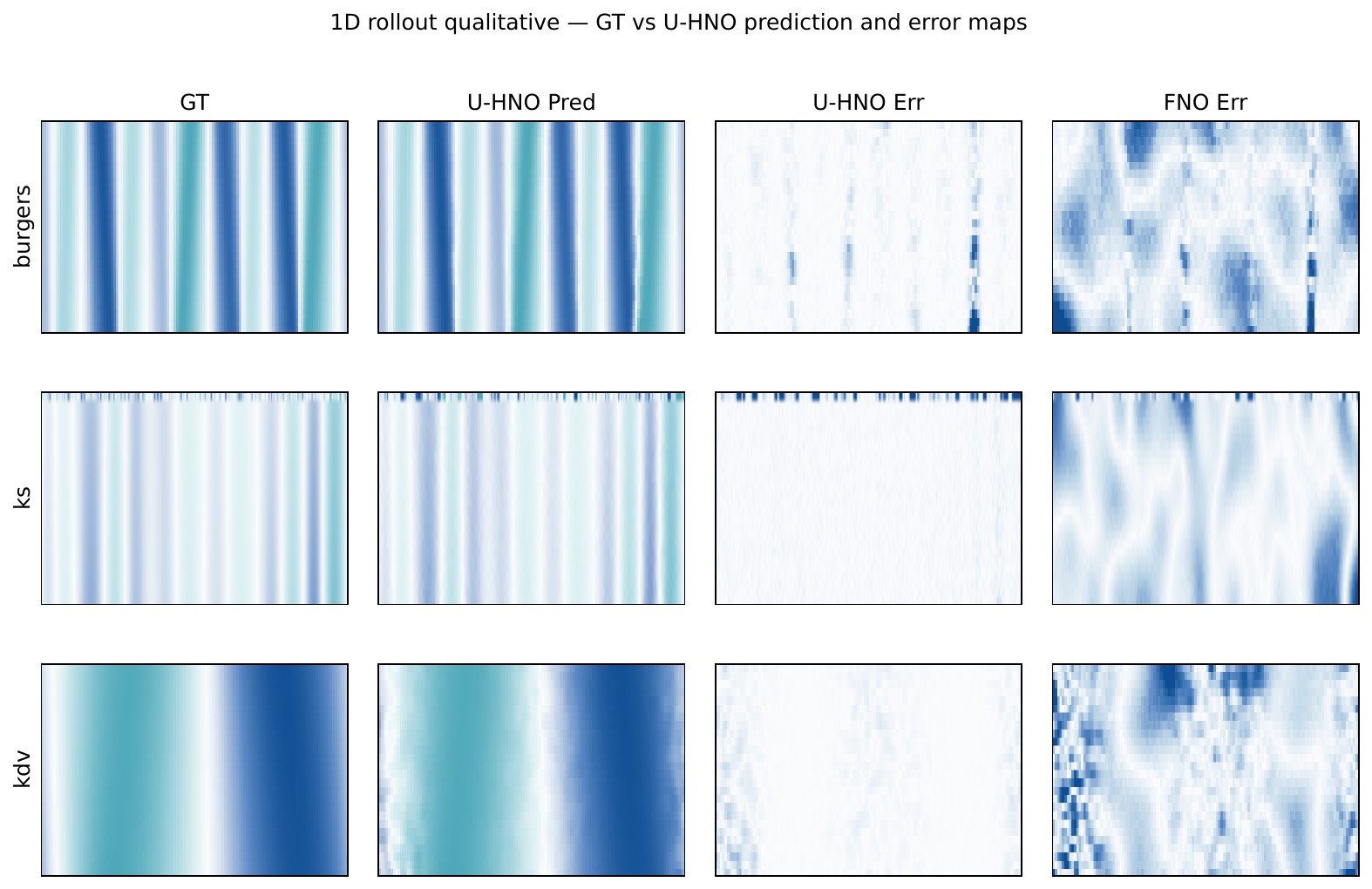}
  \caption{1D rollout space--time plots on Burgers, KS, and KdV.
  Columns: ground truth, U-HNO prediction, U-HNO absolute error,
  FNO absolute error. Errors share the same colormap range per row.}
  \label{fig:qualitative-1d}
\end{figure}

\begin{figure}[!htbp]
  \centering
  \includegraphics[width=\linewidth]{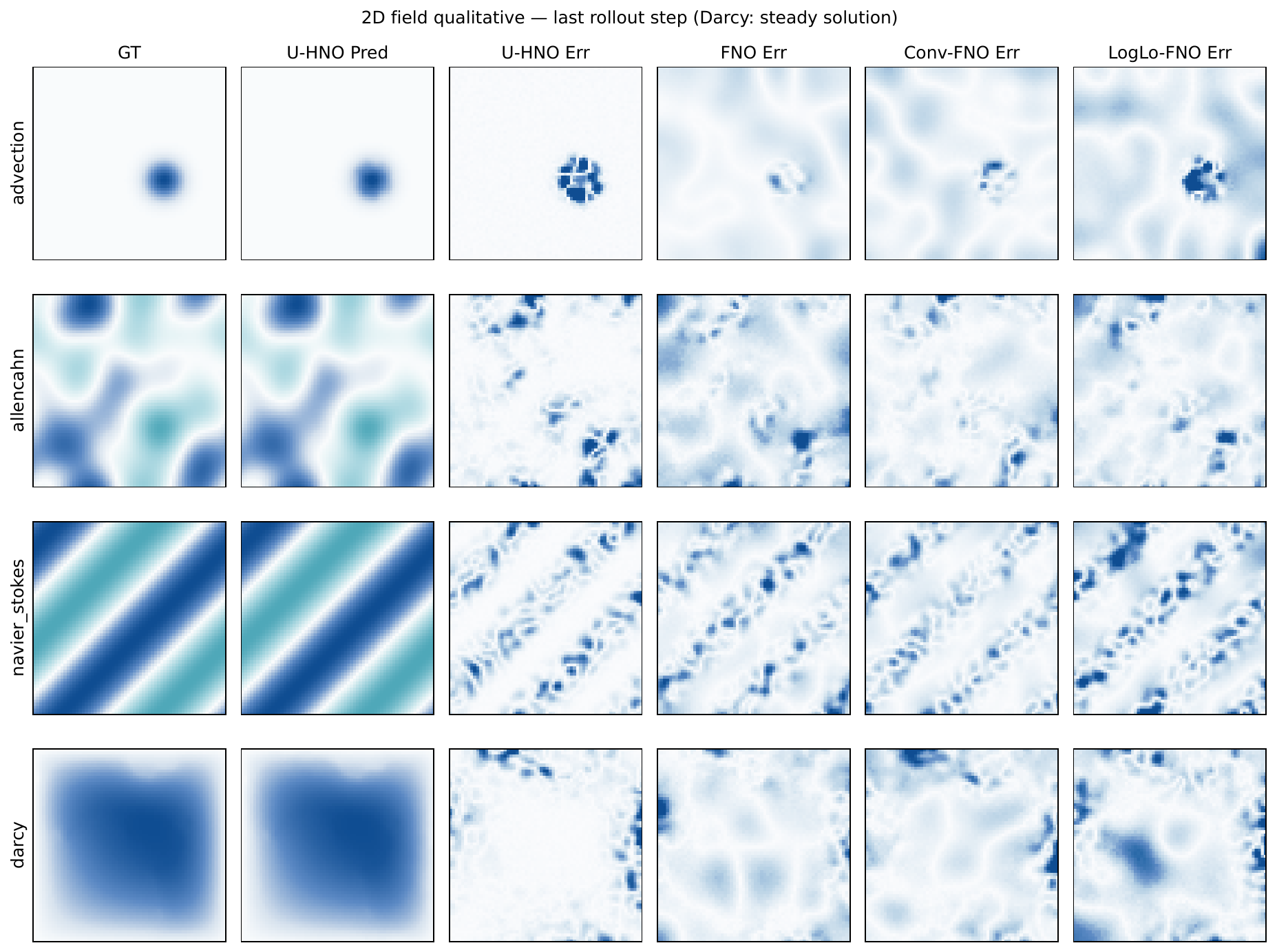}
  \caption{2D field-level qualitative plots. Columns: ground truth,
  U-HNO prediction, U-HNO error, FNO error, Conv-FNO error,
  LogLo-FNO error. Rows correspond to different PDEs.}
  \label{fig:qualitative-2d}
\end{figure}

\begin{figure}[!htbp]
  \centering
  \includegraphics[width=\linewidth]{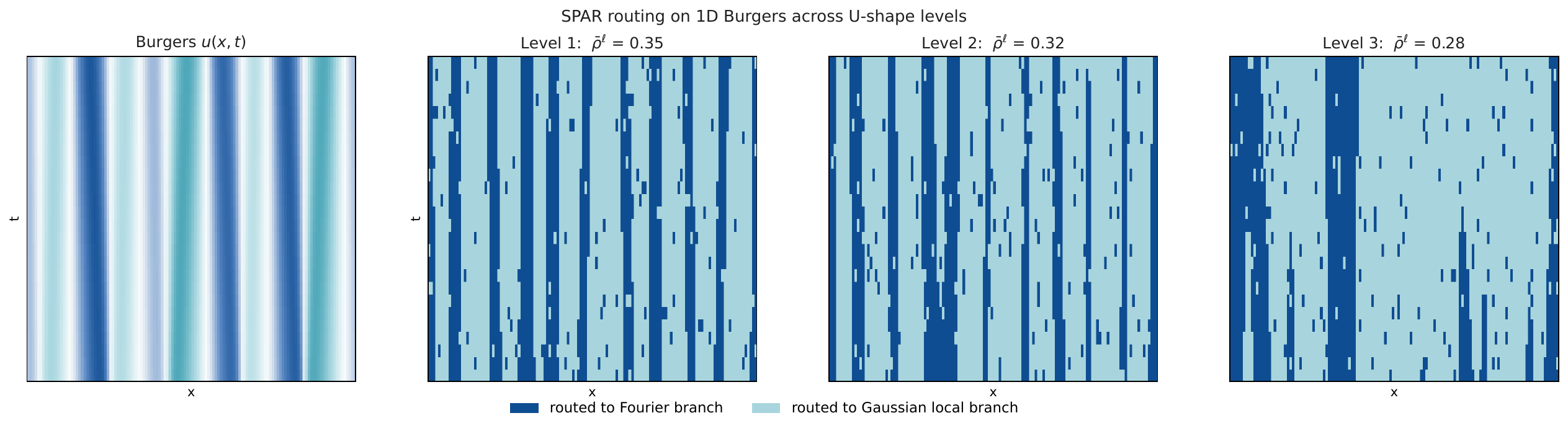}
  \caption{SPAR routing hard masks $g^{\ell}(x)$ for U-HNO on Burgers
  at every U-shape level. Red overlay indicates pixels routed to the
  Fourier branch (high $g^{\ell}$); shock fronts attract the local
  Gaussian branch (low $g^{\ell}$).}
  \label{fig:routing}
\end{figure}

\section{Evaluation Metrics}
\label{app:metrics}

We report a broad metric set because a single scalar error does not
adequately describe PDE rollout quality; the metric definitions
below are gathered for reference.

\paragraph{Relative $L^2$.}
$\operatorname{relL2}(\hat u,u)=\norm{\hat u-u}_2/\norm{u}_2$. The
standard PDE-surrogate error metric. For time-dependent tasks we
average over all rollout steps; for Darcy we report the
single-prediction value.

\paragraph{Relative $H^1$.}
\begin{equation}
\operatorname{relH1}(\hat u,u)=\frac{\bigl(\norm{\hat u-u}_2^2+\norm{\nabla\hat u-\nabla u}_2^2\bigr)^{1/2}}{\bigl(\norm{u}_2^2+\norm{\nabla u}_2^2\bigr)^{1/2}}.
\end{equation}
Captures both magnitude and gradient fidelity. The gradient
$\nabla\hat u$ is computed by central finite differences at the
prediction grid (forward-difference at the boundaries on
non-periodic Darcy). $\operatorname{relH1}$ is more discriminative
than $\operatorname{relL2}$ for shock-dominated tasks: a model can
predict the right field magnitude while smearing the gradient
profile, and only $\operatorname{relH1}$ catches that.

\paragraph{Binned spectral error.}
Partition the spatial spectrum into wavenumber-radius shells
$\{\mathcal{K}_b\}_{b=1}^{B}$; we use $B=3$ thirds (low, mid, high).
For each band,
\begin{equation}
E_b(\hat u,u)=\frac{\sum_{k\in\mathcal{K}_b}\abs{\hat U(k)-U(k)}^2}{\sum_{k\in\mathcal{K}_b}\abs{U(k)}^2},
\end{equation}
where $U=\fwd(u)$ and $\hat U=\fwd(\hat u)$. Reporting $E_b$
separately diagnoses whether a model preserves dominant low modes at
the cost of high-band content (a typical FNO failure mode) or
vice-versa.

\paragraph{Structure-function error.}
For lag $r$,
$S_r(u)=\mathbb{E}_{x}\abs{u(x+r)-u(x)}^2$
is the second-order structure function, a classical
turbulence-statistics quantity. We report a single scalar SF score
per (model, task), averaged over a log-spaced lag grid:
\begin{equation}
\mathrm{SF}(\hat u,u)=\frac{\sum_{r\in\mathcal{R}}\abs{S_r(\hat u)-S_r(u)}}{\sum_{r\in\mathcal{R}}S_r(u)},
\quad \mathcal{R}=\{1,2,4,8,16,32\}\ \text{grid units}.
\label{eq:sf}
\end{equation}
Structure-function error is sensitive to local increments and is the
natural diagnostic for whether the local branch is being recruited
where it matters.

\paragraph{Energy drift.}
\begin{equation}
\operatorname{ED}(\hat u,u)=\frac{\bigl\lvert\,\norm{\hat u_T}_2^2-\norm{u_T}_2^2\,\bigr\rvert}{\norm{u_T}_2^2}
\label{eq:ed}
\end{equation}
at the test horizon $T$. Energy drift quantifies whether the rollout
preserves the field's total $\ell_2$ energy; uncontrolled drift is a
common precursor of trajectory crash.

\paragraph{Crash rate.}
The fraction of test trajectories whose rollout diverges. We mark a
trajectory as crashed when any of the following occurs at any
rollout step: (i) NaN or $\pm\infty$ in the predicted field;
(ii) $\norm{\hat u_t}_2/\norm{u_0}_2>10^3$;
(iii) $\operatorname{relL2}(\hat u_t, u_t)>10$.

\paragraph{Rollout MSE.}
$\operatorname{MSE}=\tfrac{1}{T}\sum_{t=1}^{T}\norm{\hat u_t-u_t}_2^2$
averaged over all rollout steps and trajectories, reported for
direct comparability with prior operator-learning papers. Unlike
$\operatorname{relL2}$, $\operatorname{MSE}$ does not normalize by
ground-truth energy, so cross-task comparison requires care.

\subsection{Bootstrap Confidence Intervals}
\label{app:bootstrap}
For the main accuracy tables (Tables~\ref{tab:main-acc},
\ref{tab:main-acc-h1}) and the long-horizon stability table
(Table~\ref{tab:stability}), we report 95\% percentile bootstrap
confidence intervals over the test trajectory set ($B{=}10{,}000$
resamples). Crash rates are accompanied by Wilson 95\% intervals.
All intervals are computed at fixed model parameters from a single
training seed (seed~0); they capture variability due to the finite
test set rather than training randomness.

\end{document}